\documentclass[journal]{IEEEtran}

\ifCLASSINFOpdf
\usepackage[pdftex]{graphicx}
\graphicspath{{./}}

\DeclareGraphicsExtensions{.png,.pdf,.jpg,.jpeg}
\else
   \usepackage[dvips]{graphicx}
   \graphicspath{{Figures/}}
     \DeclareGraphicsExtensions{.png,.pdf,.jpg,.jpeg}
\fi

\usepackage{arydshln} 
\usepackage{fancyhdr}  

\usepackage{algorithmic}
\usepackage{algorithm}
\usepackage{array}
\usepackage[caption=false,font=normalsize,labelfont=sf,textfont=sf]{subfig}
\usepackage{textcomp}
\usepackage{stfloats}
\usepackage{url}
\usepackage{verbatim}
\usepackage{cite}
\usepackage{multirow}
\usepackage{pifont}
\usepackage{amsmath,amssymb,amsfonts,amsbsy}
\usepackage{bm} 
\usepackage{graphicx}
\usepackage{mathtools}
\usepackage{array,ragged2e}
\usepackage{booktabs}
\usepackage{makecell}

\usepackage{color}
\usepackage[table]{xcolor}
\definecolor{myPink}{rgb}{0.9294, 0.0078, 0.5490}
\definecolor{Gray}{gray}{0.92}
\definecolor{my_color}{HTML}{E8F3F1}
\definecolor{lightgray}{gray}{0.9}  
\usepackage[
  colorlinks=true,
  linkcolor=red,  
  citecolor=green,  
  urlcolor=myPink,   
  linkbordercolor={1 1 1} 
]{hyperref}
\usepackage{threeparttable}
\usepackage{diagbox}
\usepackage{orcidlink}

\makeatletter
\let\NAT@parse\undefined
\makeatother

\begin{document}

\title{UniUIR: Considering Underwater Image Restoration \\as An All-in-One Learner}


\author{
        Xu~Zhang,
        Huan~Zhang,
        Guoli~Wang,          
        Qian~Zhang, 
        Lefei~Zhang,~\IEEEmembership{Senior Member,~IEEE,}  
        and Bo~Du,~\IEEEmembership{Senior Member,~IEEE}


\thanks{This work was supported by the National Natural Science Foundation of China under Grants 62431020, 62225113, and the Fundamental Research Funds for the Central Universities under Grant 2042025kf0030. The numerical calculations in this work had been supported by the supercomputing system in the Supercomputing Center of Wuhan University. \emph{(Corresponding author: Lefei Zhang.)}}

\thanks{Xu Zhang, Lefei Zhang, and Bo Du are with the National Engineering
Research Center for Multimedia Software, School of Computer Science,
Wuhan University, Wuhan 430072, China (e-mail: zhangx0802@whu.edu.cn; zhanglefei@whu.edu.cn; dubo@whu.edu.cn).}


\thanks{Huan Zhang is with the School of Information Engineering, Guangdong University of Technology, Guangzhou 510006, China (e-mail: huanzhang2021@gdut.edu.cn).}

\thanks{Guoli Wang and Qian Zhang are with the Horizon Robotics, Beijing 100083, China (e-mail: guoli.wang@horizon.cc; qian01.zhang@horizon.ai).}


}

\markboth{Journal of \LaTeX\ Class Files,~Vol.~14, No.~8, August~2021}%
{Shell \MakeLowercase{\textit{\textit{et al.}}}: A Sample Article Using IEEEtran.cls for IEEE Journals}

\maketitle

\begin{abstract} 
Existing underwater image restoration (UIR) methods generally only handle color distortion or jointly address color and haze issues, but they often overlook the more complex degradations that can occur in underwater scenes. To address this limitation, we propose a \underline{Uni}versal \underline{U}nderwater \underline{I}mage \underline{R}estoration method, termed as UniUIR, considering the complex scenario of real-world underwater mixed distortions as an all-in-one manner. 
To disentangle degradation-specific effects and capture their inter-correlations, we propose the Mamba Mixture-of-Experts module (MMoEM). Each expert specializes in distinct aspects of degradation, while gating mechanism dynamically routes features to appropriate experts. This design enables collaborative prior extraction and preserves global context, all within linear computational complexity.
Building upon this foundation, to enhance degradation representation and address the task conflicts that arise when handling multiple types of degradation, we introduce the spatial-frequency prior generator. This module extracts degradation prior information in both spatial and frequency domains, and adaptively selects the most appropriate task-specific prompts based on image content, thereby improving the accuracy of image restoration. Finally, to more effectively address complex, region-dependent distortions in UIR task, we incorporate depth information derived from a large-scale pre-trained depth prediction model, thereby enabling the network to perceive and leverage depth variations across different image regions to handle localized degradation. Extensive experiments demonstrate that UniUIR can produce more attractive results across qualitative and quantitative comparisons, and shows strong generalization than state-of-the-art methods. Project page at \href{https://house-yuyu.github.io/UniUIR}{\textit{UniUIR}}.

\end{abstract}

\IEEEpeerreviewmaketitle

\section{Introduction}
\label{sec:intro} 
\IEEEPARstart{U}{nderwater} images have proliferated across fields such as marine biology, underwater exploration, and underwater archaeology\cite{cong2021rrnet}. They play a vital role in tasks like water body classification and ocean condition monitoring\cite{li2019nested}, and in providing immersive visuals of underwater heritage and landscapes.
However, due to wavelength-dependent light attenuation and scattering by marine microorganisms\cite{schettini2010underwater}, underwater scenes often exhibit multiple overlapping distortions, including haze, color shifts, low contrast, and blurred details. In addition, certain adverse imaging conditions, such as low-light environments, can further exacerbate the degradation of underwater image quality. These complex, mixed distortions pose significant challenges for both machine vision tasks and human perception, underscoring the need for effective underwater image restoration (UIR) task. 

\begin{figure}[!tp]  
	\centerline{\includegraphics[page=1,trim = 0mm 0mm 0mm 0mm, clip, width=0.98\linewidth]{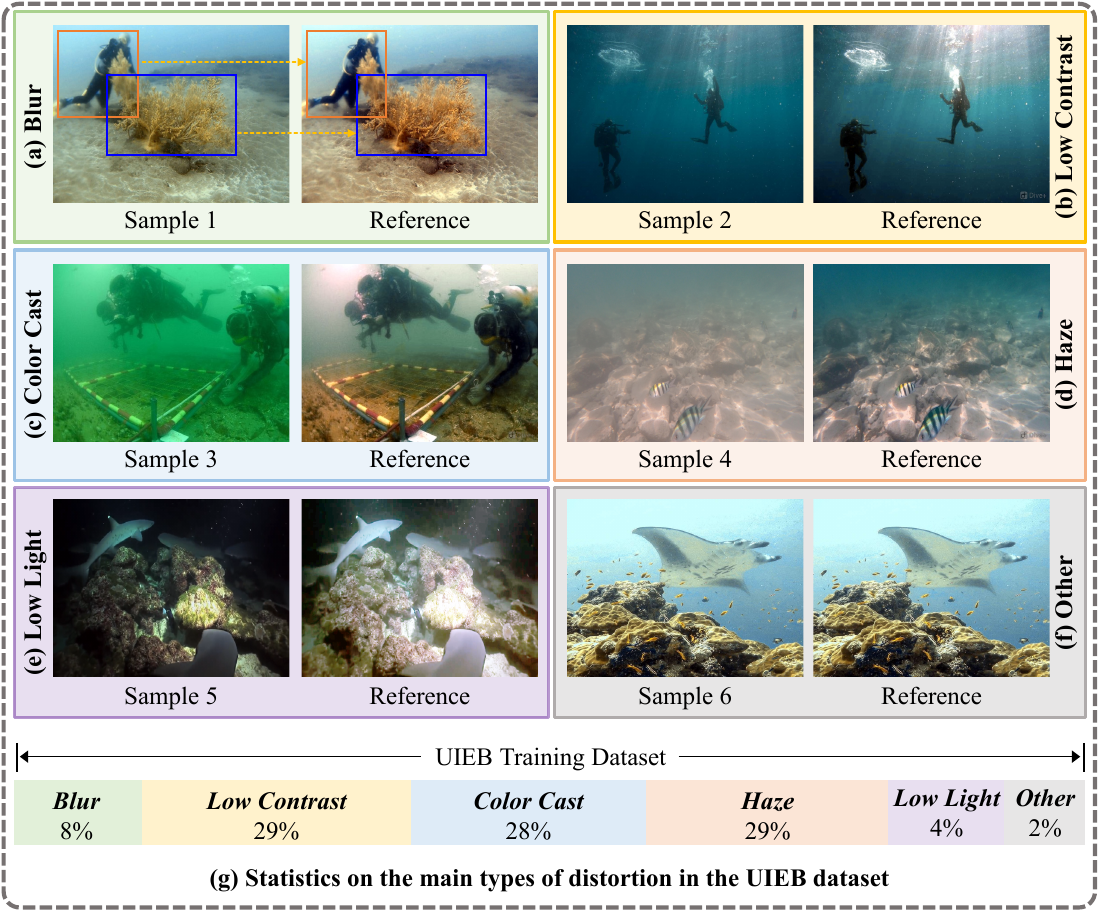}} 
 \captionsetup{skip=0pt}
	\caption{The figures present a subjective statistical analysis of the predominant distortions in the UIEB \cite{UIEB} dataset. Although each image may exhibit multiple distortions, for simplified classification, each is categorized by its most visually prominent distortion. For example, while \textit{Sample 1} shows both color distortion and foreground blurring (blue box) as well as regional blurring in the background (orange box), it is primarily classified as blurred because the blurring is more pronounced than the color distortion, which is less severe compared to other samples like \textit{Sample 3}. Reference images from the UIEB dataset are provided alongside each sample for comparison.}
	\label{fig:motivation1}
	\vspace{-1.5em}
\end{figure}

Due to the absorption of light by water, color distortion is present to varying degrees in every underwater image captured. However, in real-world scenarios, addressing only color distortion is insufficient. As shown in Fig. \ref{fig:motivation1}, we organized a group of experienced volunteers to conduct a subjective classification of distortions within the UIEB \cite{UIEB} dataset. We categorized the primary distortions in each underwater scene image into five major types: color cast, low contrast, haze, low light, and blur. It is important to note that when an image is primarily classified under one type of distortion, such as blur, it does not mean that only this type of distortion is present. For instance, as indicated by the orange and blue boxes in Sample 1 of Fig. \ref{fig:motivation1}, an image predominantly affected by blur may also exhibit color distortion in foreground objects. This highlights the extensive variety of distortion types found in underwater scenes, where multiple distortions often co-occur. Given that the underwater environment poses increasing challenges, this observation underscores the urgent need to develop comprehensive underwater restoration methods capable of handling multiple types of distortions simultaneously.

\begin{figure}[!tp]  
	\centerline{\includegraphics[page=1,trim = 0mm 0mm 0mm 0mm, clip, width=0.98\linewidth]{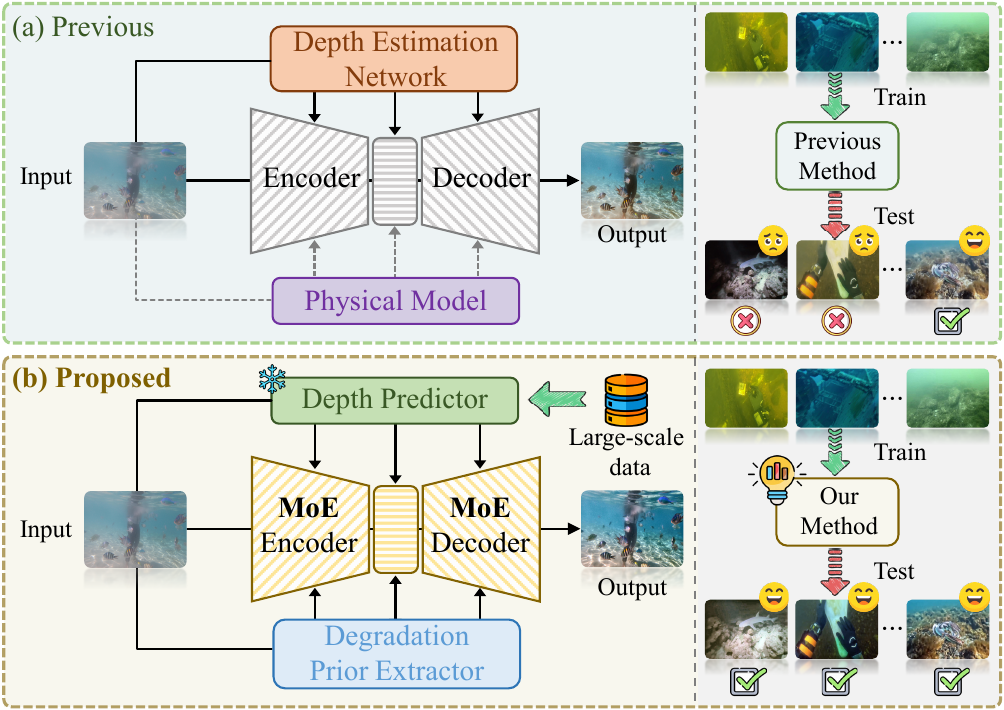}}
 \captionsetup{skip=0pt}
	\caption{Comparison between our proposed UniUIR and previous deep learning-based approaches to underwater image restoration. (a) Previous depth estimation methods rely on costly paired ground-truth depth data, while physical model-based approaches fail in complex distortions due to idealized assumptions, leading to poor generalization. (b) Our method leverages a depth prediction network trained on large-scale datasets and a reliable degradation prior extractor, demonstrating improved generalization in complex distortion scenarios through a mixture-of-experts (MoE)-based network.}
	\label{fig:motivation2}
	\vspace{-2em}
\end{figure}

Existing UIR methods can be broadly classified into three categories: visual prior-based, physical model-based, and deep learning-based approaches. Visual prior-based methods focus on adjusting image pixel values such as contrast, brightness, and saturation~\cite{ancuti2012enhancing}, but fail to account for the physical degradation processes specific to underwater environments. Physical model-based methods~\cite{li2016underwater, PM_2} mainly estimate physical medium parameters of the underwater imaging process, yet their adaptability is limited by environmental complexity, resulting in reduced robustness. 
Recently, deep learning-based approaches \cite{Ucolor, Ushape, PUGAN, NU2Net, zhou_IJCV_HCLR} have emerged, demonstrating notable improvements through end-to-end optimization. However, these methods often assume that all distortions can be handled simultaneously, typically focusing on specific types of distortions while neglecting the interactions and conflicts between them. As a result, their performance is limited in complex scenes with multiple simultaneous distortions.
As illustrated in Fig. \ref{fig:motivation2}(a), many of these methods rely on physical modeling \cite{PM_2, GUPDM} or jointly optimized depth estimation networks~\cite{PUGAN, zhou_IJCV_ADPP}, which present several challenges: 1) Depth estimation networks require paired ground-truth depth data for optimization, which is difficult and costly to obtain.  
2) Physical models rely on idealized assumptions, making them potentially inadequate for handling the complexity of real-world underwater environments, often fail to handle distortions such as detail blurring or low-light conditions. In contrast, our method (Fig. \ref{fig:motivation2}(b)) emphasizes representing degradation priors and leverages the capabilities of large-scale pre-trained depth prediction models. By incorporating an encoder-decoder framework equipped with a mixture-of-experts, our approach effectively addresses multiple distortions, including extreme low-light underwater scenarios, demonstrating robust and comprehensive restoration performance.

In this paper, we aim to systematically address the aforementioned challenges. Specifically, we propose a novel universal underwater image restoration method, termed UniUIR, designed to tackle the complex real-world scenario of mixed underwater distortions in an all-in-one manner. By analyzing the distortion characteristics and distributions in existing underwater image datasets, we develop MMoE-UIR, a Mamba Mixture-of-Experts-based restoration backbone that effectively captures and addresses diverse degradation patterns.
To further enhance restoration performance, we design a spatial-frequency prior generator (SFPG) capable of extracting degradation priors from both spatial and frequency domains while adaptively selecting the most relevant task-specific prompts based on image content. Additionally, to address inconsistencies between distortions across different image regions, we incorporate scene depth information using the Depth Anything V2 \cite{Depth_V2} model, enabling the network to better distinguish and focus on visual discrepancies between foreground and background elements.
Finally, to recover fine-grained details and mitigate information loss inherent in degradation priors, we leverage the powerful generative capabilities of diffusion models \cite{DDIM,DDPM}. The latent condition diffusion model is introduced to refine the restoration process, effectively modeling complex distributions and enabling the reconstruction of high-quality images with accurate and visually consistent details.

In summary, our main contributions are as follows.  

\noindent \ding{113}~We are among the first to address the complex scenario of underwater mixed distortions from an integrated perspective. To tackle this challenge, we propose UniUIR, a novel all-in-one method for underwater image restoration, designed to achieve more authentic restoration results.

\noindent \ding{113}~To leverage interdependencies among underwater degradations, we developed a Mamba Mixture-of-Experts module, which dynamically routes features to experts specializing in different degradation aspects and fuses their outputs to infer task-specific priors.

\noindent \ding{113}~We leverage the large-scale pre-trained depth prediction model to incorporate scene depth information, enabling the identification of distortion inconsistencies across different image regions. Additionally, we introduce the spatial-frequency prior generator, which adaptively selects the
most appropriate task-specific prompts to extract degradation priors.

\noindent \ding{113}~Compared to state-of-the-art underwater image restoration methods, UniUIR achieves superior results in extensive qualitative and quantitative evaluations. Furthermore, UniUIR demonstrates strong generalization capabilities across other visual tasks.



\section{Related Work}
\subsection{Underwater Image Restoration} 
Existing underwater image restoration (UIR) methods can be broadly divided into traditional and deep learning-based approaches. Traditional methods are further categorized into non-physical model-based and physical model-based techniques. Non-physical model-based methods focus on enhancing image quality by adjusting pixel properties such as contrast and saturation~\cite{HLRP-UIE, MLLE, ancuti2012enhancing}, without considering the optical transmission process. While computationally efficient and straightforward, these methods often lack a physical understanding of underwater light propagation, leading to issues such as over- or under-saturation in complex scenes. Physical model-based methods, on the other hand, rely on mathematical formulations of the degradation process, accounting for factors like light scattering and medium transmission. While effective in controlled settings, their reliance on predefined models often limits their adaptability to diverse underwater environments, reducing robustness and generalizability. 

Recently, deep learning based methods \cite{UIEB, Ucolor, Ushape, NU2Net, Semi-UIR, GUPDM, zhou_IJCV_ADPP, zhou_IJCV_HCLR, zhang_TCSVT_PCFB, zzk_PR} have shown significant potential in underwater image restoration (UIR). These methods leverage advanced neural network architectures and techniques to address challenges in image degradation. For instance, Waternet \cite{UIEB} integrates enhanced input modes with confidence maps to generate improved images, while Ucolor \cite{Ucolor} employs multi-color space embedding guided by medium transmission to enhance detail and color restoration. Semi-UIR \cite{Semi-UIR} combines contrastive and semi-supervised learning to improve performance and robustness, and HCLR-net \cite{zhou_IJCV_HCLR} uses hybrid contrastive learning with locally randomized perturbations to enhance image quality. Despite these advancements, deep learning-based methods often assume that all distortions can be addressed simultaneously. This assumption leads to a focus on main distortion, such as color cast or low contrast, while overlooking interactions and conflicts between multiple distortions. Consequently, their performance remains limited in complex underwater scenes with diverse and simultaneous degradations.

\subsection{All-in-One Image Restoration} 
All-in-one image restoration \cite{AirNet, PromptIR, ProRes, PerceiveIR, MPerceiver} has emerged as a promising approach in low-level vision tasks, aiming to recover clean images from multiple degraded inputs using a unified model. Compared to task-specific \cite{AdcSR, CS_TIP, MoRE, AFDformer, S3Net, Diff-Retinex, huan_eswa, WNN_1, zlf_tcb, WNN_2, zlf_cvpr} and general \cite{Restormer, MambaIR, MDDA-former, DiffIR} restoration methods, all-in-one restoration offers advantages in model storage efficiency and practical applications. The main challenge lies in designing a single architecture capable of addressing diverse degradation types. To tackle this, methods like AirNet \cite{AirNet} employ contrastive learning for discriminative degradation representations. Vision prompts, as used in PromptIR \cite{PromptIR} and ProRes \cite{ProRes}, further enhance the handling of diverse degradations. More recently, MPerceiver \cite{MPerceiver} and Perceive-IR\cite{PerceiveIR} have leveraged pre-trained large-scale vision models to excel in all-in-one
restoration tasks.

Underwater Image Restoration (UIR) naturally aligns with the all-in-one restoration framework due to the presence of multiple distortions in underwater scenes (as shown in Fig. \ref{fig:motivation1}). However, directly applying general all-in-one methods to underwater images is suboptimal because of the significant domain shift between natural and underwater environments. To address this gap, we propose a specialized all-in-one restoration approach tailored for UIR. By introducing a mixture-of-experts mechanism, our method adopts a ``divide-and-conquer" strategy, dynamically allocating expert submodules to handle specific types of distortions, thereby improving restoration precision and adaptability. Additionally, our method integrates physical depth information and emphasizes spatial and frequency domain prior representations, enabling it to better capture the unique characteristics of underwater imagery. These innovations allow our approach to achieve superior restoration results, particularly in challenging underwater environments with complex and overlapping distortions.

\subsection{Diffusion Models for Image Restoration}
Diffusion models (DMs) \cite{DDIM, DDPM}, originally developed for generative tasks, have demonstrated strong potential in image restoration by addressing inverse problems through progressive denoising. Their capability to model complex data distributions has led to state-of-the-art results in low-level vision tasks such as super-resolution \cite{SD_SR, SR-Diff}. Recently, the generative power of DM has attracted attention in underwater image restoration (UIR) \cite{DM-Water, UIEDP}. For example, Tang et al. \cite{DM-Water} proposed a diffusion-based underwater image enhancement method, while UIEDP \cite{UIEDP} introduced a framework incorporating diffusion priors, formulating UIR as posterior sampling conditioned on degraded inputs. 

While DMs offer a principled way to model uncertainty, they still face practical limitations in UIR, including low sampling efficiency, poor adaptability to diverse degradation types, insufficient integration of environmental priors (e.g., depth, scattering properties), and limited generalization.
To address these issues, we propose UniUIR, a unified framework built upon a novel Latent Conditional Diffusion Model (LCDM). Unlike standard DMs, LCDM explicitly incorporates high-level semantic and structural priors into the denoising process, enabling degradation-aware, fine-grained restoration. By conditioning on these priors in the latent space, our model achieves enhanced perceptual quality and structural coherence, while improving efficiency and generalization through targeted guidance.

\begin{figure*}[!htp]  
	\centerline{\includegraphics[page=1,trim = 0mm 0mm 0mm 0mm, clip, width=1\linewidth]{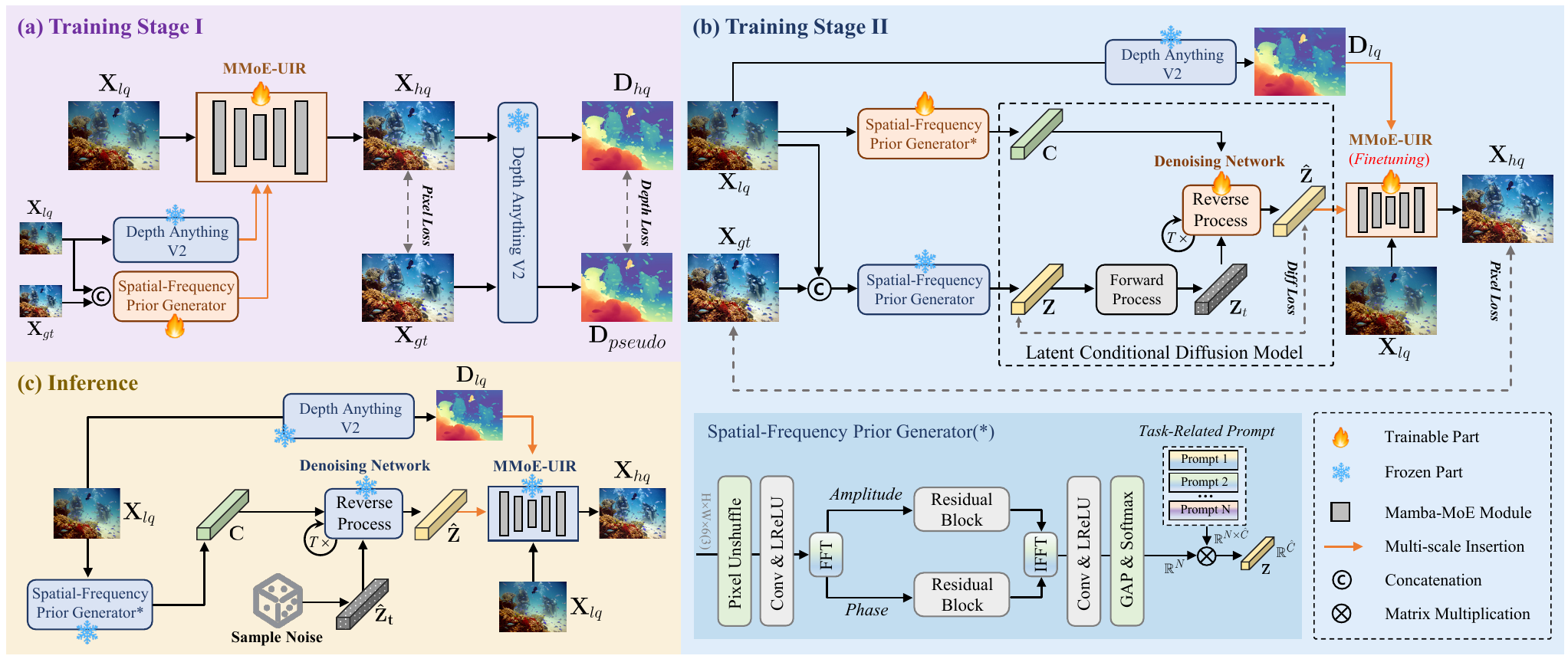}}
  \captionsetup{skip=0pt}
	\caption{Overview of the proposed UniUIR. (a) In \textbf{Stage I}, the depth map $\mathbf{D}_{lq}$, degradation prior $\mathbf{Z}$, and image $\mathbf{X}_{lq}$ are processed by the MMoE-UIR network to reconstruct the high-quality image $\mathbf{X}_{hq}$. The $\mathcal{L}_{1}$ loss and the edge-aware depth loss $\mathcal{L}_{\text{depth}}$ encourage the network to focus on overall pixel-level reconstruction and the accurate recovery of edges and structures, respectively. (b) In \textbf{Stage II}, the prior $\mathbf{Z}$ extracted by the SFPG and conditional embedding $\mathbf{C}$ will undergoes Latent Condition Diffusion Model to iteratively denoise, yielding $\mathbf{\hat{Z}}$. Then, $\mathbf{D}_{lq}$, $\mathbf{X}_{lq}$, and $\mathbf{\hat{Z}}$ are processed by MMoE-UIR to restore the clean image. The network is fine-tuned with $\mathcal{L}_{1}$ and $\mathcal{L}_{\text{diff}}$ to optimize image restoration and the diffusion process. (c) During \textbf{Inference}, SFPG* first produces the conditional embedding \(\mathbf{C}\). Subsequently, a random noise \(\mathbf{\hat{Z}_{t}}\) is sampled from a Gaussian distribution. Through the reverse process, \(\mathbf{C}\) and \(\mathbf{\hat{Z}_{t}}\) undergo iterative denoising to generate \(\mathbf{\hat{Z}}\). Lastly, the fine-tuned MMoE-UIR then leverages $\mathbf{X}_{lq}$, $\hat{\mathbf{Z}}$, and $\mathbf{D}_{lq}$ to reconstruct the final high-quality image $\mathbf{X}_{hq}$.}  
	\label{fig:UniUIR}
	\vspace{-1.5em}
\end{figure*}

\section{Method}
\label{Method}

\subsection{Overview Pipeline} 
The proposed UniUIR framework comprises three stages: training stage I, training stage II, and inference stage. As illustrated in Fig. \ref{fig:UniUIR}, during training stage I, given an underwater image $\mathbf{X}_{lq} \in \mathbb{R}^{H \times W \times 3}$, a pre-trained Depth Anything V2 model is used to incorporate scene depth information $\mathbf{D}_{lq}$ from $\mathbf{X}_{lq}$. Meanwhile, the reference image $\mathbf{X}_{gt}$ is concatenated with $\mathbf{X}_{lq}$ and passed through the spatial-frequency prior generator (SFPG) to obtain a degradation prior $\mathbf{Z} \in \mathbb{R}^{\hat{C}}$. Subsequently, $\mathbf{D}_{lq}$ and $\mathbf{Z}$ together with $\mathbf{X}_{lq}$ are processed by the MMoE-UIR, an $n$-stage U-shaped network, to reconstruct high-quality result $\mathbf{X}_{hq}$. 
In training stage II, the prior \(\mathbf{Z}\) extracted by the pre-trained SFPG undergoes a forward diffusion process to generate the noised \(\mathbf{Z}_t\), and the \(\mathbf{X}_{lq}\) is passed through SFPG* to yield a conditional embedding \(\mathbf{C}\). During the reverse process, \(\mathbf{Z}_t\) and \(\mathbf{C}\) are fed into a UNet–based denoising network to iteratively perform denoising operations, yielding \(\mathbf{\hat{Z}}\). Finally, \(\mathbf{D}_{lq}\), \(\mathbf{X}_{lq}\), and \(\mathbf{\hat{Z}}\) are processed by MMoE-UIR to restore a clean underwater image. Throughout this process, MMoE-UIR is fine-tuned based on the initial training stage. The optimization employs both pixel loss to constrain the overall image restoration and Diff loss to regulate the diffusion model's generation process.
During inference stage, given $\mathbf{X}_{lq}$, SFPG* first produces the conditional embedding \(\mathbf{C}\). Subsequently, a random noise \(\mathbf{\hat{Z}_{t}}\) is sampled from a Gaussian distribution. Through the reverse process, \(\mathbf{C}\) and \(\mathbf{\hat{Z}_{t}}\) undergo iterative denoising to generate \(\mathbf{\hat{Z}}\). Lastly, the fine-tuned MMoE-UIR then leverages $\mathbf{X}_{lq}$, $\hat{\mathbf{Z}}$, and the extracted depth $\mathbf{D}_{lq}$ to reconstruct the final high-quality image $\mathbf{X}_{hq}$.
In the following subsections, we will provide a detailed description of the roles and processes of these modules.

\noindent\textbf{Optimization.}
In the first stage, we use $\mathcal{L}_{1}$ loss as the pixel-level loss to constrain the overall learning process, in addition, we introduce edge-aware depth loss $\mathcal{L}_{\text{depth}}$ to let the network focus on the gradient information of the depth map, emphasizing the accurate recovery of edges and structures.
The training loss can be describe as follows:
\begin{equation}
\mathcal{L}_{\text{stageI}} = \mathcal{L}_{{1}}(\mathbf{X}_{gt}, \mathbf{X}_{hq})+\lambda_{1}\mathcal{L}_{\text{depth}},  
\end{equation}
\begin{equation}
\mathcal{L}_{\text{depth}} = \mathcal{L}_{1}(\mathbf{D}_{pseudo}, \mathbf{D}_{hq})+\lambda_{2}\mathcal{L}_{\text{grad}}(\mathbf{D}_{pseudo}, \mathbf{D}_{hq}),  
\end{equation}
\begin{equation}
\mathcal{L}_{\text{grad}} =
\left( \left| \frac{\partial({\mathbf{D}_{{pseudo}}-\mathbf{D}_{hq}})}{\partial x} \right|
+ \left| \frac{\partial({\mathbf{D}_{{pseudo}}-\mathbf{D}_{hq}})}{\partial y} \right| \right),  
\end{equation}
where $\mathbf{X}_{gt}$ and $\mathbf{X}_{hq}$ denote the ground-truth image and the restored high-quality image, respectively. Similarly, $\mathbf{D}_{pseudo}$ and $\mathbf{{D}}_{hq}$ represent the pseudo depth and high-quality depth maps derived from the Depth Anything V2 Model using $\mathbf{X}_{gt}$ and $\mathbf{X}_{hq}$, respectively. The hyperparameters $\lambda_{1}$ and $\lambda_{2}$ are introduced to balance the contributions of individual terms in the total loss function. Our training strategy incorporates both edge-aware depth loss and $\mathcal{L}_{1}$ loss to achieve complementary benefits. The edge-aware depth loss enables the model to capture precise edge information from the pseudo ground-truth depth, while the $\mathcal{L}_{1}$ loss ensures accurate overall depth estimation. This dual-loss approach enhances depth prediction performance by effectively combining edge precision with global consistency.

In the second stage, we jointly optimize using the diffusion loss $\mathcal{L}_{\text{diff}}$ and pixel-level loss. The training loss can be described as follows:
\begin{equation}
\mathcal{L}_{\text{stageII}} = \mathcal{L}_{{1}}(\mathbf{X}_{gt}, \mathbf{X}_{hq})+\mathcal{L}_{\text{diff}}, \enspace\mathcal{L}_{\text{diff}} = \mathcal{L}_{{1}}(\mathbf{Z}, \mathbf{\hat{Z}}). 
\end{equation}

\subsection{MMoE-UIR} 
\begin{figure*}[!tp]  
	\centerline{\includegraphics[page=1,trim = 0mm 0mm 0mm 0mm, clip, width=0.99\linewidth]{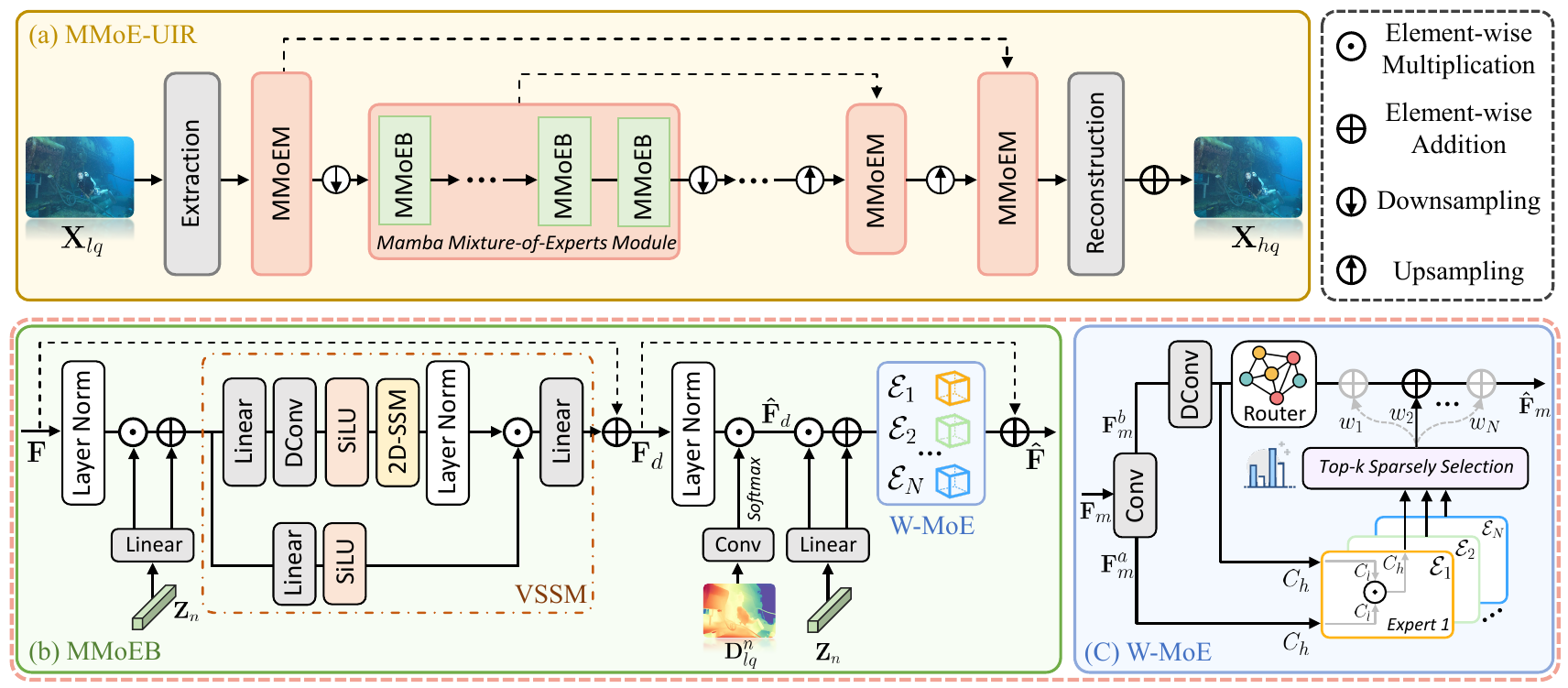}}
  \captionsetup{skip=0pt}
	\caption{The detailed structure of the proposed MMoE-UIR. (a) Mamba Mixture-of-Experts Underwater Image Restoration (MMoE-UIR); (b) Mamba Mixture-of-Experts Block (MMoEB); (c) Water Mixture-of-Experts (W-MoE). }  
	\label{fig:MMoE-UIR}
	\vspace{-1.5em}
\end{figure*}

As depicted in Fig. \ref{fig:MMoE-UIR}, the low-quality underwater input \(\mathbf{X}_{lq}\) is processed by a $3\times3$ convolution-based feature extraction module, followed by a four-stage U-shaped encoder–decoder network. The core component, the Mamba Mixture of Experts Block (MMoEB), integrates the Vision State-Space Module (VSSM) and Water Mixture-of-Experts (W-MoE). Each MMoEM contains multiple MMoEBs. By leveraging task-specific degradation priors and depth information, the MMoEB refines these priors to facilitate multi-scale feature fusion and reconstruction, improving adaptability to complex underwater environments and overall restoration performance.

\noindent\textbf{Mamba Mixture-of-Experts Block.}
As shown in Fig. \ref{fig:MMoE-UIR}(b), given the deep feature $\mathbf{F}\in \mathbb{R}^{H \times W \times C}$, we first use the Layer Norm (LN) followed. Similarly, we allow the prior feature \(\mathbf{z}\) to interact with the normalized feature before feeding it into the VSSM. This process can be represented as:
\begin{equation}
\mathbf{F}_d=\text{VSSM}\big(\text{LN}(\mathbf{F})\odot\text{Linear}(\mathbf{Z}_n)+\text{Linear}(\mathbf{Z}_n)\big)+\mathbf{F},  
\end{equation}
{where $\mathbf{Z}_n$ represents the $n$-th stage compression feature, which is obtained by downsampling the original feature $\mathbf{Z}$ for $n$ times.} The generated deep features $\mathbf{F}_d$ are subsequently fed into the FFN model, where they are first enhanced through LN. The normalized features then interact with the enhanced deep features, after which they are passed into the W-MoE. This process can be expressed as:
\begin{equation}
\mathbf{\hat{F}}_d=\text{LN}(\mathbf{F}_{d})\odot\text{Softmax}\big( \mathcal{C}_{3\times3}(\mathbf{D}_{lq}^n)\big),
\end{equation}
\begin{equation}
\mathbf{\hat{F}}=\mathcal{M}\big(\mathbf{\hat{F}}_d\odot\text{Linear}(\mathbf{Z}_n)+\text{Linear}(\mathbf{Z}_n)\big)+\mathbf{F}_d,
\end{equation}
where $\mathbf{\hat F}$ represents the output of the MMoEB; $\mathcal{M}(\cdot)$ denotes the Water Mixture-of-Experts operation; $\mathcal{C}_{3\times3}$ denotes 3$\times$3 convolution, followed by a ReLU function; $\mathbf{D}_{lq}$ is generated from the pre-trained Depth Anything V2 model. Each MMoEM contains multiple MMoEBs.

\noindent\textbf{Water Mixture-of-Experts.} 
As illustrated in Fig. \ref{fig:MMoE-UIR}(c), given an input feature $\mathbf{F}_{m}\in \mathbb{R}^{H \times W \times C}$ to W-MoE, we employ a $3\times3$ convolution for feature projection and subsequently split the features along the channel dimension to create two distinct views $\mathbf{F}_{m}^{a}$ and $\mathbf{F}_{m}^{b}\in \mathbb{R}^{H \times W \times C}$. Simultaneously, a parallel depth-wise convolution extracts the local spatial context $\mathbf{\hat{F}}_{m}^{a}$ before feeding both extracted feature maps into the Mixture-of-Experts. To further investigate the intricacies of inter-dependencies among the extracted features while reducing complexity, we implement low-rank decomposition for the inputs while modeling global contextual relationships. A single low-rank expert $\mathcal{E}_{i}$ can be formulated as:
\begin{equation}
    \mathcal{E}_i = \mathcal{T}_{C_l \rightarrow C_h}^3 \left( \mathcal{T}_{C_h \rightarrow C_l}^1 (\mathbf{{F}}_{m}^a) \odot \mathcal{T}_{C_h \rightarrow C_l}^2 (\mathbf{\hat{F}}_{m}^b) \right),
\end{equation}
here, for the $i$-th expert $\mathcal{E}_{i}$, the number of channels $C_h, C_l, C$ are related as follows: $C_h=C$ and $C_l=2^{i+1}$. The linear transfer layer, denoted as $\mathcal{T}(\cdot)$, is implemented using $1\times1$ point-wise convolutions that compress the encoded features along the channel dimension to their low-rank approximation $R_i$, where $i \in {1, \cdots, N}$. After modulating spatial cues through element-wise multiplication with contextual cues in the low-dimensional space, another linear transfer layer, $\mathcal{T}_{C_l \rightarrow C_h}^3$, restores the features back to their original dimension $C$. This reconstruction enables the effective extraction of channel-wise spatial content while implicitly integrating critical spatial and channel dependencies.
Additionally, we employ a routing network $\mathcal{G}$, composed of two MLP layers. This network systematically explores the search space to identify the optimal low-rank expert based on the input features and the network depth. The final output, $\mathbf{\hat{F}}_{m}$, of the W-MoE is formulated as follows:
\begin{equation}
    \mathbf{\hat{F}}_{m} = \sum_{i=1}^{N} \mathcal{G}(\mathbf{{\hat F}}_{m}^{b})\cdot \mathcal{E}_i(\mathbf{{F}}_{m}^{a}, \mathbf{{\hat F}}_{m}^{b}),
\end{equation}
where $\mathcal{G}(\cdot)$ represents the learned router module, while $\mathcal{E}_{i}(\cdot)$ denotes the output of the $i$-th expert. The sparsity inherent in the router function $\mathcal{G}(\cdot)$ optimizes computational efficiency by assigning higher weights to the top-$k$ low-rank experts. During training, the model learns from all experts; during inference, only the selected top-$k$ experts are utilized for computation, significantly enhancing efficiency. For clarity, the pseudocode of the W-MoE is provided in Algorithm \ref{alg:moe_layer}.

\subsection{Spatial-Frequency Prior Generator} 
Reconstructing image details in severely degraded regions is challenging due to the significant loss of high-frequency information caused by complex distortions. To restore these missing details, external guidance from high-quality underwater images is necessary. To achieve this, we use a Spatial-Frequency Prior Generator (SFPG) to extract a compact prior $\mathbf{Z}$ from $\mathbf{X}_{lq}$ and $\mathbf{X}_{gt}$, injecting it into MMoE-UIR to provide correct
information for restoring the lost details.
Specifically, as shown in Fig. \ref{fig:UniUIR}, SFPG first employs PixelUnshuffle and 3$\times$3 convolutions with LReLU activation functions to downsample the concatenated images $\mathbf{X}_{c} \in \mathbb{R}^{H\times W \times6}$ and capture effective information, resulting in $\mathbf{X}_{s} \in \mathbb{R}^{\frac{H}{4}\times\frac{W}{4} \times C'}$. Subsequently, two parallel branches of three residual blocks with 3$\times$3 convolutions and LReLU activation functions are utilized to extract features from the amplitude and phase of $\mathbf{X}_{s}$.
\begin{equation}
\mathbf{X}_\mathcal{A}, \mathbf{X}_\mathcal{P} = \mathcal{F}(\mathbf{X}_{s}), \enspace\tilde{\mathbf{X}}_\mathcal{A}, \tilde{\mathbf{X}}_\mathcal{P} = \text{ResBlock}(\mathbf{X}_\mathcal{A}, \mathbf{X}_\mathcal{P}),
\end{equation}
\begin{equation}
\mathbf{X}_{f} = \mathcal{F}^{-1}(\tilde{\mathbf{X}}_\mathcal{A}, \tilde{\mathbf{X}}_\mathcal{P}),
\end{equation}
where \(\mathcal{F}(\cdot)\) denotes the 2D FFT and \(\mathcal{F}^{-1}(\cdot)\) represents the 2D IFFT, \({\mathbf{X}}_\mathcal{A}\) and \({\mathbf{X}}_\mathcal{P}\) respectively denote the amplitude and phase features. After being mapped back to the spatial domain, these features are refined by three convolutional layers with LReLU activation functions and a global average pooling (GAP), producing a compact representation.

To effectively accommodate various degradation types of underwater image restoration tasks within a unified
model, we embed Task-Related Prompts (TRP) at the end of the SFPG, generating prompt information closely tied to specific tasks. Concretely, we employ the softmax output \(\mathcal{S} \in \mathbb{R}^{N}\) as the weighting factor for the basic prompts associated with each degradation, to combine and construct
prompts for that particular task. Let \(\mathcal{P} \in \mathbb{R}^{N \times \hat{C}}\) be the basic task-related prompts. The final generated prior \(\mathbf{Z}\) can be expressed as:
\begin{equation}
\mathbf{Z} = {\mathcal{P}} \otimes \mathcal{S},
\end{equation}
where $\otimes$ represents matrix multiplication.

\begin{algorithm}[t] 
    \caption{Water Mixture-of-Experts}
    \label{alg:moe_layer}
\begin{algorithmic}[1]
    \STATE \textbf{Input:} Input feature $\mathbf{{F}}_{m}$, separated feature $\mathbf{{F}}_{m}^{a}$, extracted local features $\mathbf{{\hat F}}_{m}^{b}$
    \STATE \textbf{Parameters:} $N$ Experts $\mathcal{E}$, Router $\mathcal{G}$, and top-k expert
    
    \STATE Compute router outputs: $\mathbf{g} = \mathcal{G}(\mathbf{\hat{x}}_{a})$
    \STATE Normalize weights: $\mathbf{w} = \text{Softmax}(\mathbf{g})$
    \STATE Select top-k expert: $w_{1}, w_{2},...,w_{k}= \text{top-k}(\mathbf{w}, k)$
    
    \IF{training}
        \FOR{each $\mathcal{E}_{i} \in \mathcal{E}$}
            \STATE $\mathbf{y}^{i}_{\mathcal{E}_{i}} = \mathcal{T}_{C_l \rightarrow C_h}^3 \left( \mathcal{T}_{C_h \rightarrow C_l}^1 (\mathbf{{F}}_{m}^a) \odot \mathcal{T}_{C_h \rightarrow C_l}^2 (\mathbf{\hat{F}}_{m}^b) \right)$
        \ENDFOR
        \STATE Compute final output: $\mathbf{y} = \sum_{i=1}^{N} w_i \cdot \mathbf{y}^{i}_{\mathcal{E}_{i}}$
    \ELSE
        \STATE Compute final output: $\mathbf{y} = w_{\text{top-k}} \cdot \mathbf{y}^{\text{top-k}}_{\mathcal{E}_{i}}$
    \ENDIF
    
    \STATE \textbf{Output:} Final output $\mathbf{y}$
\end{algorithmic}
\end{algorithm}

\subsection{Latent Conditional Diffusion Model} 
\noindent\textbf{Forward Diffusion Process.} 
As illustrated in Fig. \ref{fig:MMoE-UIR}(b), we first utilize the SFPG trained in the first stage to generate \( \mathbf{Z} \). Subsequently, a diffusion process is applied to gradually add Gaussian noise to \( \mathbf{Z} \) over \( T \) time steps, resulting in the noisy representation \( \mathbf{Z}_T \), which can be expressed as:
\begin{equation}
q(\mathbf{Z}_T \mid \mathbf{Z}) = \mathcal{N}(\mathbf{Z}_T; \sqrt{\bar{\alpha}_T} \mathbf{Z}, (1 - \bar{\alpha}_T) \mathbf{I}),   
\end{equation}
where $T$ is total number of iterations; $\alpha_T = 1 - \beta_T$ and $\bar{\alpha}_T = \prod_{i=1}^{T} \alpha_i$. ${\beta}_t (t=1,\cdots,T)$ is hyper parameter controlling the variance of the noise; $\mathcal{N}$ denotes the Gaussian distribution.


\noindent\textbf{Reverse Diffusion Process.} 
The reverse process runs backwards from $\mathbf{Z}_T$ to $\mathbf{\hat Z}$. For the
reverse step from $\mathbf{Z}_t$ to $\mathbf{Z}_{t-1}$, we use the posterior distribution as:
\begin{equation}
p(\mathbf{Z}_{t-1} \mid \mathbf{Z}_t, \mathbf{\hat Z}) = \mathcal{N}(\mathbf{Z}_{t-1}; \boldsymbol{\mu}_t(\mathbf{Z}_t, \mathbf{\hat Z}), \frac{1 - \bar{\alpha}_{t-1}}{1 - \bar{\alpha}_t} \beta_t \mathbf{I}), 
\end{equation}  %
\begin{equation}\label{eq: epsilon}
\boldsymbol{\mu}_t(\mathbf{Z}_t, \mathbf{\hat Z}) = \frac{1}{\sqrt{\alpha_t}} (\mathbf{Z}_t - \frac{1 - \alpha_t}{\sqrt{1 - \bar{\alpha}_t}} \boldsymbol{\epsilon}), 
\end{equation} 
where 
$\epsilon$ represents the noise in $\mathbf{Z}_t$. We adopt a UNet as denoising network to estimate the noise $\epsilon$ for each step. We utilize another spatial-frequency prior generator, denoted as SFPG$ \ast $, which maintains the same structure as SFPG, except that it operates without Pixel Unshuffle and accepts only $\mathbf{X}_{lq}$ as input. SFPG$ \ast $ compresses the underwater image into condition latent feature $\mathbf{C} \in \mathbb{R}^{N\times \hat{C}}$. 
The denoising network predicts the noise conditioned on the $\mathbf{Z}_t$ and $\mathbf{C}$, i.e., ${\epsilon}_{\theta}(\mathbf{Z}_t, \mathbf{C}, t)$. With the substitution of ${\epsilon}_{\theta}$ in Eq. (\ref{eq: epsilon}) and set the variance to $(1 - \alpha_t)$, we can get:
\begin{equation}\label{eq: denoise}
\mathbf{Z}_{t-1} = \frac{1}{\sqrt{\alpha_t}} \left( \mathbf{Z}_t - \frac{1 - \alpha_t}{\sqrt{1 - \bar{\alpha}_t}} \boldsymbol{\epsilon}_\theta(\mathbf{Z}_t, \mathbf{C}, t) \right) + \sqrt{1 - \alpha_t} \boldsymbol{\epsilon}_t,
\end{equation}
where ${\epsilon}_t\sim\mathcal{N} (0, \mathbf{I})$. By iteratively sampling $\mathbf{Z}_t$ using Eq. (\ref{eq: denoise}) $T$ times, we can generate the predicted prior feature $\mathbf{\hat{Z}} \in \mathbb{R}^{N\times \hat{C}}$. The predicted prior feature is then used to guide MMoEB in Fig. \ref{fig:MMoE-UIR}(b). 
\begin{table*}[tbp]
\setlength{\abovecaptionskip}{2pt}
\caption{Datasets summary for underwater image restoration task.}
\label{tab:dataset}
\centering
  \renewcommand\arraystretch{1.2} 
    \resizebox{\linewidth}{!}{\begin{tabular}{c|c|c|c|c|c|c}
    
\toprule[1pt]
\toprule[0.5pt]

{\bf Setting} &\bf Train/Test & {\bf Dataset} & {\bf Type} & {\bf No. of Images} & {\bf Main Objects} & \bf Resolution \\ 

\hline

{\multirow{7}{*}{Set \scalebox{1.15}{\ding{182}}}}
&\multirow{1}{*}{{{Training}}}  
& {UIEB-T\cite{UIEB}} 
& Paired
& {800} 
& Wreck, People, Fish, Reef, and Coral
& (259$\times$194), $\cdots$, (2180$\times$1447)
\\  

\cdashline{2-7}

{}
&\multirow{6}{*}{{{Testing}}}  
& T90\cite{UIEB} 
& Paired
& 90 
& Wreck, People, Fish, Reef, and Coral
& (500$\times$333), $\cdots$, (1200$\times$850)
\\ 

{}
&
& U45\cite{U45}
& Unpaired
& 45
& Wreck, People, Fish, Reef, and Coral
& 256$\times$256
\\ 

{}
&
& SQUID-16\cite{SQUID}
& Unpaired
& 16
& Wreck and Reef
& 512$\times$512
\\  

{}
&
&  Challenge-60\cite{UIEB}
& Unpaired
&  60
&  People, Fish, Reef, Medusa
&  (259$\times$194), $\cdots$, (2000$\times$1124)
\\  

{}
&
&  UCCS\cite{UCCS}
& Unpaired
&  300
&  Reef and Sea Urchin
&  400$\times$300
\\ 

{}
&
&  EUVP-330\cite{EUVP}
& Unpaired
& 330
&  People, Fish, Reef, and Coral
& (320$\times$240), $\cdots$, (960$\times$540)
\\ 
\hline

{\multirow{2}{*}{Set \scalebox{1.15}{\ding{183}}}}
&\multirow{1}{*}{{{Training}}}
& {LSUI-T\cite{Ushape}} 
& Paired
& {3879} 
& Fish, Reef, Coral, and Sea Urchin
& (256$\times$256), $\cdots$, (1280$\times$1024)
\\  
\cdashline{2-7}

{}
&{\multirow{1}{*}{{{Testing}}}}
& {LSUI-400\cite{Ushape}} 
& Paired
& 400
& Fish, Reef, Coral, and Sea Urchin
& (256$\times$256), $\cdots$, (1280$\times$1024)
\\  

    \bottomrule[1pt]
    \end{tabular}}
    \vspace{-1.5em}
\end{table*}

\section{Experiments}
\subsection{Experimental Setup} 
\noindent\textbf{Datasets.} 
As shown in Tab. \ref{tab:dataset}, the datasets used in this study include both paired and unpaired sets. For paired training data, we utilize UIEB-T \cite{UIEB} and LSUI-T \cite{Ushape}. Unpaired test datasets include U45 \cite{U45}, SQUID-16 \cite{SQUID}, Challenge-60 \cite{UIEB}, UCCS \cite{UCCS}, and EUVP-330 \cite{EUVP}. The UIEB dataset consists of 950 real underwater images, 890 of which have corresponding reference images. We use 800 of these paired images for training (UIEB-T) and the remaining 90 for testing (T90). The remaining 60 images, lacking satisfactory reference images, constitute the Challenge-60 subset. The LSUI dataset is a large-scale underwater image dataset comprising 4279 image pairs. Following \cite{Ushape}, we use 3879 images for training (LSUI-T) and the remaining 400 for testing (LSUI-400). The U45 dataset presents 45 real-world underwater images with challenges including color casts, low contrast, and haze-like effects. The SQUID dataset contains 57 stereo underwater image pairs, each including a color chart for color restoration evaluation. Following \cite{Ucolor}, we selected 16 representative samples from SQUID, referred to as SQUID-16. Further details about the datasets are provided in Tab. \ref{tab:dataset}.

\noindent\textbf{Evaluation Metrics.} 
We evaluate performance using reference metrics including Peak Signal-to-Noise Ratio (PSNR), Structural Similarity (SSIM), and Learned Perceptual Image Patch Similarity (LPIPS) \cite{LPIPS}, as well as non-reference metrics such as the Underwater Colour Image Quality Evaluation Metric (UCIQE) \cite{UCIQE}, Underwater Image Quality Measure (UIQM) \cite{UIQM}, URanker \cite{NU2Net}, Underwater Image Fidelity (UIF) \cite{UIF}, and Underwater Image Quality Index (UIQI) \cite{UIQI}. 
For the metrics PSNR, SSIM, UCIQE, UIQM, URanker, UIF, and UIQI, higher scores indicate better performance. Conversely, for the LPIPS metric, lower scores are preferable.

\noindent\textbf{Baselines.}
We compare our method against existing underwater image restoration methods, encompassing both traditional and deep learning approaches. These include a traditional method MLLE \cite{MLLE} and the following deep learning models: WaterNet \cite{UIEB}, FUnIE \cite{FUnIE}, S-UWNet \cite{S-UWNet}, Ucolor \cite{Ucolor}, PUIE-Net \cite{PUIE-Net}, Ushape \cite{Ushape}, PUGAN \cite{PUGAN}, NU2Net \cite{NU2Net}, Semi-UIR \cite{Semi-UIR}, GUPDM\cite{GUPDM}, and HCLR-Net \cite{zhou_IJCV_HCLR}. Additionally, we compared our approach with the all-in-one image restoration method PromptIR\cite{PromptIR} and the MambaIR\cite{MambaIR} based on the Mamba architecture \cite{Mamba}.

\noindent\textbf{Implementation Details.}  
All experiments were conducted using PyTorch on 8 NVIDIA RTX 3090 GPUs and optimized using the AdamW optimizer with momentum parameters (0.9, 0.999). During training stage I and II, UniUIR was trained for 50K and 200K iterations, respectively, with an initial learning rate of $2\times10^{-4}$, gradually reduced to $1\times10^{-6}$ using cosine annealing. The batch size was set to 8. MMoE-UIR utilized a 4-level encoder-decoder architecture with [3, 5, 6, 6] MMoEBs and channel dimensions of [32, 64, 128, 256] for levels 1 through 4. 
The channel number ${\hat C}$ for both SFPG and SFPG* was 256. 
Task-related prompts (256-dimensional) were initialized as learnable parameters. Within the W-MoE module, we employed 3 experts and activated the top-$k$ ($k$=2) for each input.
A linear noise schedule was employed for diffusion process with parameters $\alpha_{1}=0.99$, $\alpha_{T}=0.1$, and a time step of $T=4$. Depth information was extracted using the large version of the Depth Anything V2 model \cite{Depth_V2}.
For fair comparisons, all learning-based UIR methods were retrained using the original settings specified in their respective papers. Input images were cropped to 128$\times$128 pixels, and random flipping was applied for data augmentation. The hyperparameters were set as $\lambda_{1}=0.1$ and $\lambda_{2}=0.5$.

\begin{figure*}[!tbp]  
	\centerline{\includegraphics[page=1,trim = 0mm 0mm 0mm 0mm, clip, width=0.98\linewidth]{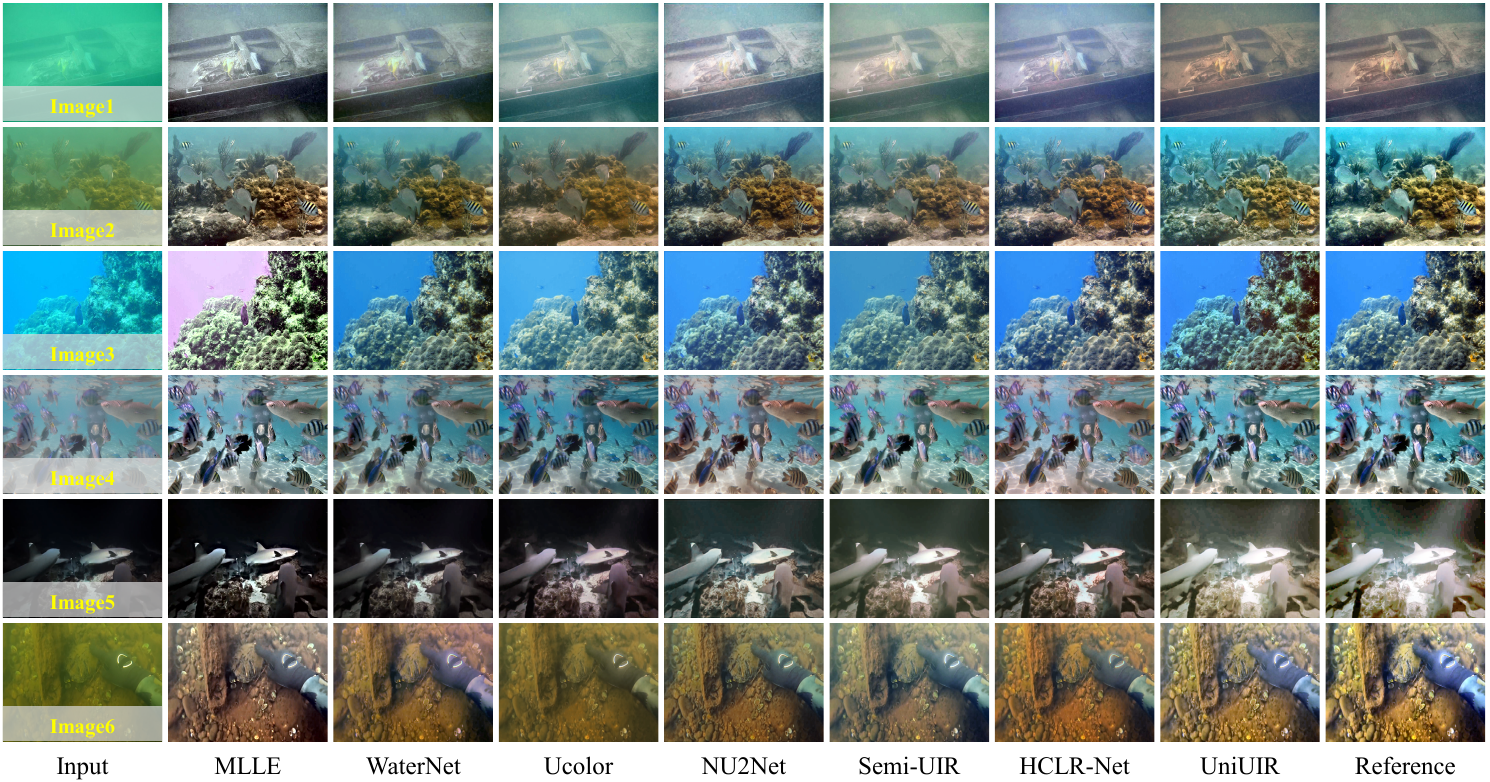}}
  \captionsetup{skip=0pt}
	\caption{Visual comparison of restored results for the T90\cite{UIEB} test set.}
	\label{fig:T90}
	\vspace{-1em}
\end{figure*}

\begin{table*}[!tbp]
\setlength{\abovecaptionskip}{2pt}
\caption{
Quantitative results on \textbf{referenced-based} datasets: T90 and LSUI-400. The top three results are marked with \textcolor{red}{{red}}, \textcolor{blue}{{blue}}, and \textcolor{green}{{green}}, respectively.}

\label{tab:ref}

\centering
\renewcommand\arraystretch{1.2}
\tabcolsep=0.1cm 
\resizebox{0.98\linewidth}{!}{\begin{tabular}{l|c|cccccc|cccccc}  

\toprule[1pt]
\toprule[0.5pt]
\multirow{2}{*}{Method}     
& \multirow{2}{*}{Source}
& \multicolumn{6}{c|}{\cellcolor{lightgray}{T90} \cite{UIEB}}   
& \multicolumn{6}{c}{\cellcolor{lightgray}{LSUI} \cite{Ushape}}                                  
\\ 
\cline{3-14} 

{} 
&
& \multicolumn{1}{c}{{PSNR$\uparrow$}} 
& \multicolumn{1}{c}{{SSIM$\uparrow$}}
& \multicolumn{1}{c}{{LPIPS$\downarrow$}}
& \multicolumn{1}{c}{{UCIQE$\uparrow$}} 
& \multicolumn{1}{c}{{UIQM$\uparrow$}}
& \multicolumn{1}{c|}{{URanker$\uparrow$}} 

& \multicolumn{1}{c}{{PSNR$\uparrow$}} 
& \multicolumn{1}{c}{{SSIM$\uparrow$}}
& \multicolumn{1}{c}{{LPIPS$\downarrow$}}
& \multicolumn{1}{c}{{UCIQE$\uparrow$}} 
& \multicolumn{1}{c}{{UIQM$\uparrow$}} 
& \multicolumn{1}{c}{{URanker$\uparrow$}} 
\\  

\midrule
WaterNet \cite{UIEB}   
& TIP'19
& 20.87& 0.911 & 0.157 & 0.595 & 2.823 & 1.371
& 23.07& 0.853 & 0.221 & 0.584 & 2.807 & 1.398
\\

FUnIE \cite{FUnIE}
& IRAL'20
& 19.18 & 0.865 & 0.179 & \textcolor{green}{0.599} & 3.002 & 1.157
& 22.05 & 0.832 & 0.239 & 0.567 & 2.681 & 1.083
\\

S-UWnet \cite{S-UWNet} 
& AAAI'21
& 18.37 & 0.753 & 0.192 & 0.581 & 2.432 & 1.298
& 21.33 & 0.884 & 0.243 & 0.591 & 2.719 & 1.385
\\

Ucolor \cite{Ucolor}
& TIP'21
& 20.81 & 0.904 & 0.160 & 0.573 & 3.019 & 1.416
& 23.28 & 0.890 & 0.211 & 0.589 & 2.804 & 1.201
\\

MLLE \cite{MLLE}  
& TIP'22
& 18.68 & 0.855 & 0.286 & 0.598 & 2.854 & 1.007
& 22.23 & 0.835 & 0.254 & 0.569 & 3.071 & 1.116
\\

PUIE-Net \cite{PUIE-Net}   
& ECCV'22
& 22.03 & 0.891 & 0.241 & 0.587 & 2.946 & 1.462
& 23.93 & 0.904 & 0.206 & 0.614 & 2.981 & 1.418
\\

Ushape \cite{Ushape} 
& TIP'23
& 20.97 & 0.864 & 0.203 & 0.579 & 3.067 & 1.358                 
& 24.46 & 0.901 & 0.190 & 0.599 & 2.994 & 1.411
\\

PUGAN \cite{PUGAN}  
& TIP'23
& 22.79 & 0.922 & 0.193 & 0.592 & \textcolor{green}{3.104} & 1.347      
& 25.11 & {0.915} & 0.184 & \textcolor{blue}{0.615} & \textcolor{green}{3.112} & 1.309
\\

NU2Net \cite{NU2Net}  
& AAAI’23
& 22.91 & 0.922 & 0.174 & \textcolor{red}{0.605}  & \textcolor{red}{3.153} & \textcolor{blue}{1.517}           
& 25.32 & 0.910 & 0.162 &    0.611  & 3.005 & 1.413
\\    

Semi-UIR \cite{Semi-UIR} 
& CVPR’23
& \textcolor{green}{24.19} & {0.923} & 0.151 & 0.568 & 2.537 & {1.479}
& {27.08} & 0.907 & \textcolor{green}{0.134} & \textcolor{green}{0.614} & 3.016 & 1.435
\\

GUPDM \cite{GUPDM}   
& ACM MM’23
&\textcolor{blue}{24.33} & \textcolor{blue}{0.928} & \textcolor{green}{0.133} &0.552 &3.012 & 1.306
&\textcolor{blue}{27.77} & \textcolor{green}{0.917} & 0.138 &\textcolor{red}{0.627} &3.103 
& {1.466}
\\

PromptIR \cite{PromptIR}   
& NeurIPS’23
&23.46 & {0.919} & 0.152 &0.571 &2.933  & 1.376
&26.69 & 0.905   & 0.141 &0.594 &2.984  & 1.321
\\

HCLR-Net \cite{zhou_IJCV_HCLR}   
& IJCV’24
& 23.72 & 0.917 & 0.158 & 0.575 & 2.537 & 1.454
& 26.98 & 0.904 & \textcolor{blue}{0.130} & 0.597 & \textcolor{red}{3.272} & \textcolor{green}{1.483}
\\

MambaIR \cite{MambaIR}   
& ECCV'24
& 23.57 & 0.922 & {0.147} &0.584 &2.851 & 1.385
& 26.32 & 0.905 & 0.143 &0.602 &3.056 & 1.322
\\

{DiffUIR \cite{DiffUIR}}   
& {CVPR'24} 
& 24.05 & 0.925 & 0.136 & 0.592 & 2.939 & 1.465 
& 27.19 & 0.911 & {0.137} & 0.605 & 3.071 & 1.474
\\

{AdaIR \cite{AdaIR}}   
& {ICLR'25}
& 24.16 & \textcolor{green}{0.926} & \textcolor{blue}{0.131} & 0.587 & 3.052 & \textcolor{green}{1.483}
& \textcolor{green}{27.55} & \textcolor{blue}{0.920} & 0.139 & 0.599 & 3.104 & \textcolor{blue}{1.485}
\\

\rowcolor{my_color}
{UniUIR (Ours)}  
& ---
& \textcolor{red}{25.11}   & \textcolor{red}{0.933}  & \textcolor{red}{0.112} & \textcolor{blue}{0.601}  
& \textcolor{blue}{3.117}  & \textcolor{red}{1.534}       
& \textcolor{red}{28.42} & \textcolor{red}{0.926} &\textcolor{red}{0.123} & {0.610}& \textcolor{blue}{3.138}
& \textcolor{red}{1.507}
\\

\bottomrule[0.8pt]
\end{tabular}}
\vspace{-1.5em}
\end{table*}

\begin{table*}[tb]
\setlength{\abovecaptionskip}{2pt}
\caption{
Quantitative results on \textbf{non-referenced-based} datasets: U45, SQUID-16, Challenge-60, UCCS, and EUVP-330. The top three results are marked with \textcolor{red}{{red}}, \textcolor{blue}{{blue}}, and \textcolor{green}{{green}}, respectively.}

\label{tab:noref}

\centering
\renewcommand\arraystretch{1.2}
\tabcolsep=0.05cm 
\resizebox{\linewidth}{!}{\begin{tabular}{l|c|ccc|ccc|ccc|ccc|ccc}  

\toprule[1pt]
\toprule[0.5pt]
\multirow{2}{*}{Method} 
& \multirow{2}{*}{Source}
& \multicolumn{3}{c|}{\cellcolor{lightgray}{U45} \cite{U45}}      
& \multicolumn{3}{c|}{\cellcolor{lightgray}{SQUID-16} \cite{SQUID}}                                  & \multicolumn{3}{c|}{\cellcolor{lightgray}{Challenge-60} \cite{UIEB}}   
& \multicolumn{3}{c|}{\cellcolor{lightgray}{UCCS} \cite{UCCS}}   
& \multicolumn{3}{c}{\cellcolor{lightgray}{EUVP-330} \cite{EUVP}}
\\ 
\cline{3-17}

{} 
&
& \multicolumn{1}{c}{UCIQE$\uparrow$} 
& \multicolumn{1}{c}{UIQM$\uparrow$}
& \multicolumn{1}{c|}{URanker$\uparrow$}

& \multicolumn{1}{c}{UCIQE$\uparrow$} 
& \multicolumn{1}{c}{UIQM$\uparrow$}
& \multicolumn{1}{c|}{URanker$\uparrow$}

& \multicolumn{1}{c}{UCIQE$\uparrow$} 
& \multicolumn{1}{c}{UIQM$\uparrow$}
& \multicolumn{1}{c|}{URanker$\uparrow$}

& \multicolumn{1}{c}{UCIQE$\uparrow$} 
& \multicolumn{1}{c}{UIQM$\uparrow$}
& \multicolumn{1}{c|}{URanker$\uparrow$}

& \multicolumn{1}{c}{UCIQE$\uparrow$} 
& \multicolumn{1}{c}{UIQM$\uparrow$}
& \multicolumn{1}{c}{URanker$\uparrow$}

\\  

\midrule
WaterNet \cite{UIEB}      
& TIP'19
& 0.572 &3.195 &1.312
& {{0.541}} &{{2.318}} &1.746
& 0.566 &2.653 &1.223
& 0.545 &3.058 &1.281 
& 0.524 &3.019 &1.685  
\\

FUnIE \cite{FUnIE} 
& IRAL'20
& {{0.589}}&{{3.181}}&1.763
& 0.532&{{2.417}}&1.456 
& 0.570&\textcolor{blue}{{3.158}}&1.304
& 0.541&2.986&1.043 
& 0.548&2.942&1.728  
\\

S-UWnet \cite{S-UWNet} 
& AAAI'21
& 0.533&2.933&1.681
& 0.491&1.894&1.279
& 0.466&2.396&1.473
& 0.486&2.759&1.312  
& 0.523&3.018&1.609 
\\

Ucolor \cite{Ucolor}   
& TIP'21
& 0.564&3.151&1.485
& 0.514&2.215&1.613
& 0.532&2.746&1.213
&0.550&3.019&1.064
&0.561&\textcolor{green}{3.114}&1.546
\\

MLLE \cite{MLLE}     
& TIP'22
& 0.593&2.599&1.349
& 0.531&2.314&1.378
& {0.581} & 2.310 & 1.257
& 0.544&2.985&1.145 
& \textcolor{green}{0.568}&3.026&1.629
\\

PUIE-Net \cite{PUIE-Net}    
& ECCV'22
& 0.578&2.799&1.632           
& 0.522&1.923&2.014
& 0.558&2.521&1.307
& 0.536&3.003&1.352 
& 0.548&2.979&1.415  
\\

Ushape \cite{Ushape}   
& TIP'23
& 0.553&3.048&1.737                  
& 0.528&2.256&1.857                  
& 0.534&2.783&1.363
& 0.567&3.012&1.374 
& 0.557&3.013&1.679 
\\

PUGAN \cite{PUGAN}     
& TIP'23
& {{0.590}}&3.095&1.815         
& \textcolor{blue}{{0.566}} & 2.399 & 1.956      
& \textcolor{red}{{0.612}} & {3.001} & 1.641
& 0.536&2.977&1.549 
& 0.541&2.858&1.747 
\\

NU2Net \cite{NU2Net}   
& AAAI’23
& \textcolor{green}{0.595}&\textcolor{green}{{3.206}}&1.804          
& 0.534&{{2.480}}&\textcolor{blue}{2.156}           
& 0.564&{{2.907}}&1.557
& \textcolor{red}{0.601}&2.994&\textcolor{blue}{1.624} 
& \textcolor{blue}{0.571}&\textcolor{blue}{3.127}&1.801  
\\   

Semi-UIR \cite{Semi-UIR}   
& CVPR’23
& \textcolor{blue}{0.601}&\textcolor{red}{3.323}&1.831
& {0.558} &\textcolor{red}{2.537}&1.945
& 0.573&2.925 &{1.685}
& 0.552&3.039 &1.321
& 0.526&2.994 &\textcolor{green}{1.864} 
\\

GUPDM \cite{GUPDM}   
& ACM MM’23
& 0.566&3.057&\textcolor{green}{1.872}
& 0.547&2.459&2.013 
& 0.548&2.687&\textcolor{blue}{1.751}
& \textcolor{blue}{0.586}&\textcolor{blue}{3.157}&1.583
& 0.563&3.014&1.733
\\

PromptIR \cite{PromptIR}   
& NeurIPS’23
&0.573 & 2.847 & 1.736 
&0.531 &2.473 & 1.834
&0.577 & 2.808 & 1.638 
&0.549 &2.897  & 1.512
&0.545 &2.911 & 1.844
\\

HCLR-Net \cite{zhou_IJCV_HCLR}   
& IJCV’24
& 0.585&3.013&\textcolor{blue}{1.927}
& 0.538&\textcolor{green}{2.508}&\textcolor{green}{2.109}
& 0.569&2.737&1.615
& 0.541&3.009&\textcolor{red}{1.784} 
& 0.559&3.102&\textcolor{red}{2.052}
\\

MambaIR \cite{MambaIR}   
& ECCV’24
& 0.589&3.127&1.825 
& 0.553&2.322&1.895
& 0.552&2.983&1.493
& 0.564&\textcolor{green}{3.114}&1.417 
& 0.537&2.975&1.806
\\

{DiffUIR \cite{DiffUIR}}   
& {CVPR'24}
& 0.593 & 3.153 & 1.819 
& 0.549 & 2.415 & 2.034
& 0.575 & \textcolor{green}{3.057} & \textcolor{green}{1.722}
& 0.567 & 3.102 & 1.598
& 0.566 & 3.023 & 1.857
\\

{AdaIR \cite{AdaIR}}   
& {ICLR'25}
& 0.582 & 3.195 & 1.852 
& \textcolor{green}{0.563} & 2.479 & 2.075
& \textcolor{green}{0.587} & 2.942 & 1.706 
& 0.559 & 3.073 & 1.564
& 0.560 & 3.105 & 1.823
\\

\rowcolor{my_color}
{UniUIR (Ours)}    
& ---
& \textcolor{red}{{0.609}}&\textcolor{blue}{3.278}&\textcolor{red}{2.028}    
& \textcolor{red}{{0.570}}&\textcolor{blue}{2.514}&\textcolor{red}{2.217}          
& \textcolor{blue}{{0.593}}&\textcolor{red}{3.218}&\textcolor{red}{1.825}
& \textcolor{green}{0.576}&\textcolor{red}{3.215}&\textcolor{green}{1.603}  
& \textcolor{red}{0.581}&\textcolor{red}{3.198}&\textcolor{blue}{1.927} 
\\

\bottomrule[0.8pt]
\end{tabular}}
\vspace{-1em}
\end{table*}

\begin{figure}[!tbp]  %
	\centerline{\includegraphics[page=1,trim = 0mm 0mm 0mm 0mm, clip, width=0.99\linewidth]{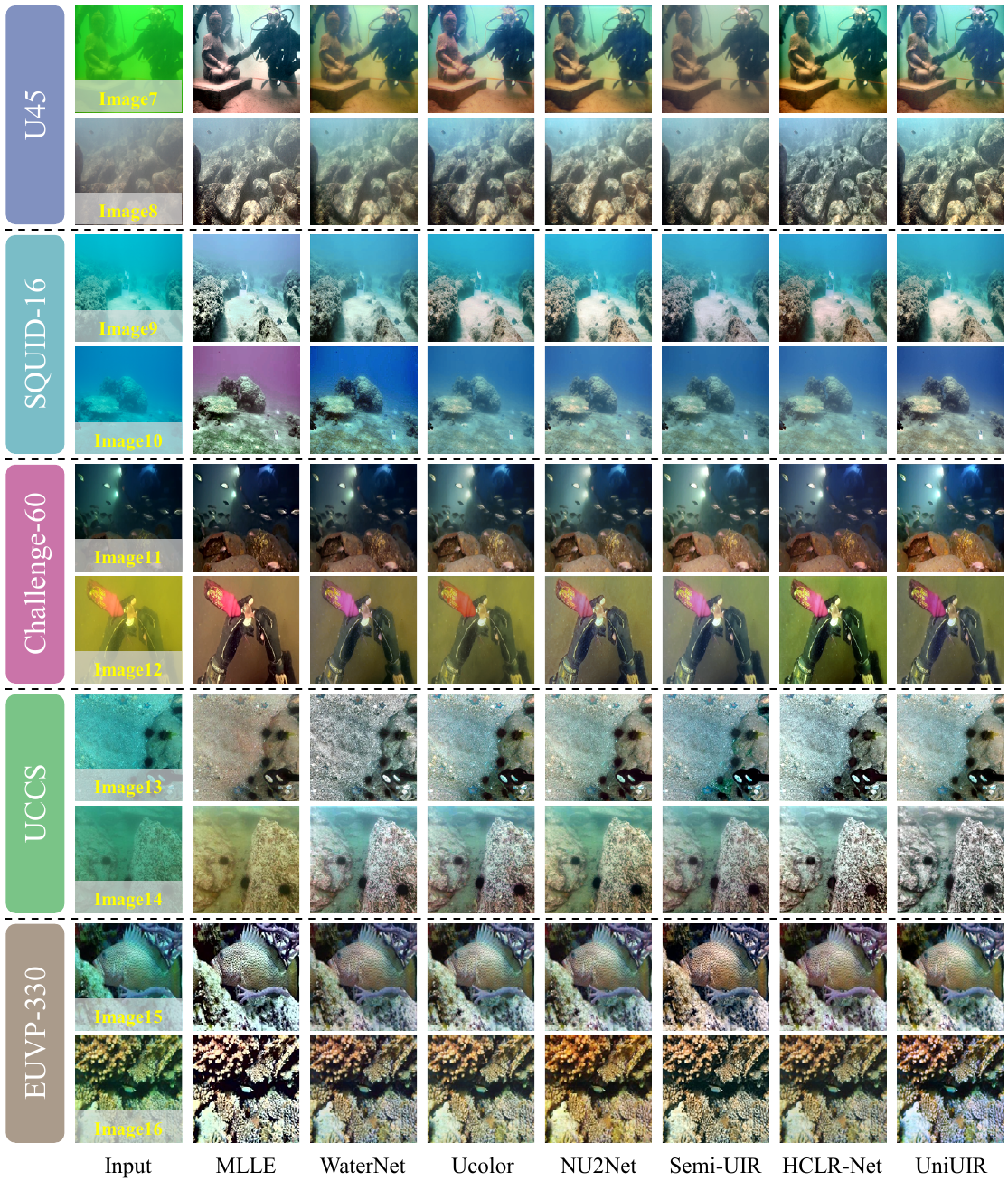}}
  \captionsetup{skip=0pt}
	\caption{Visual comparison of restoration results on the non-reference underwater image datasets: U45 \cite{U45}, SQUID-16 \cite{SQUID}, Challenge-60 \cite{UIEB}, UCCS \cite{UCCS}, and EUVP-330 \cite{EUVP}.}
	\label{fig:noref}
	\vspace{-1em}
\end{figure}

\begin{figure}[!tbp] 
	\centerline{\includegraphics[page=1,trim = 0mm 0mm 0mm 0mm, clip, width=1\linewidth]{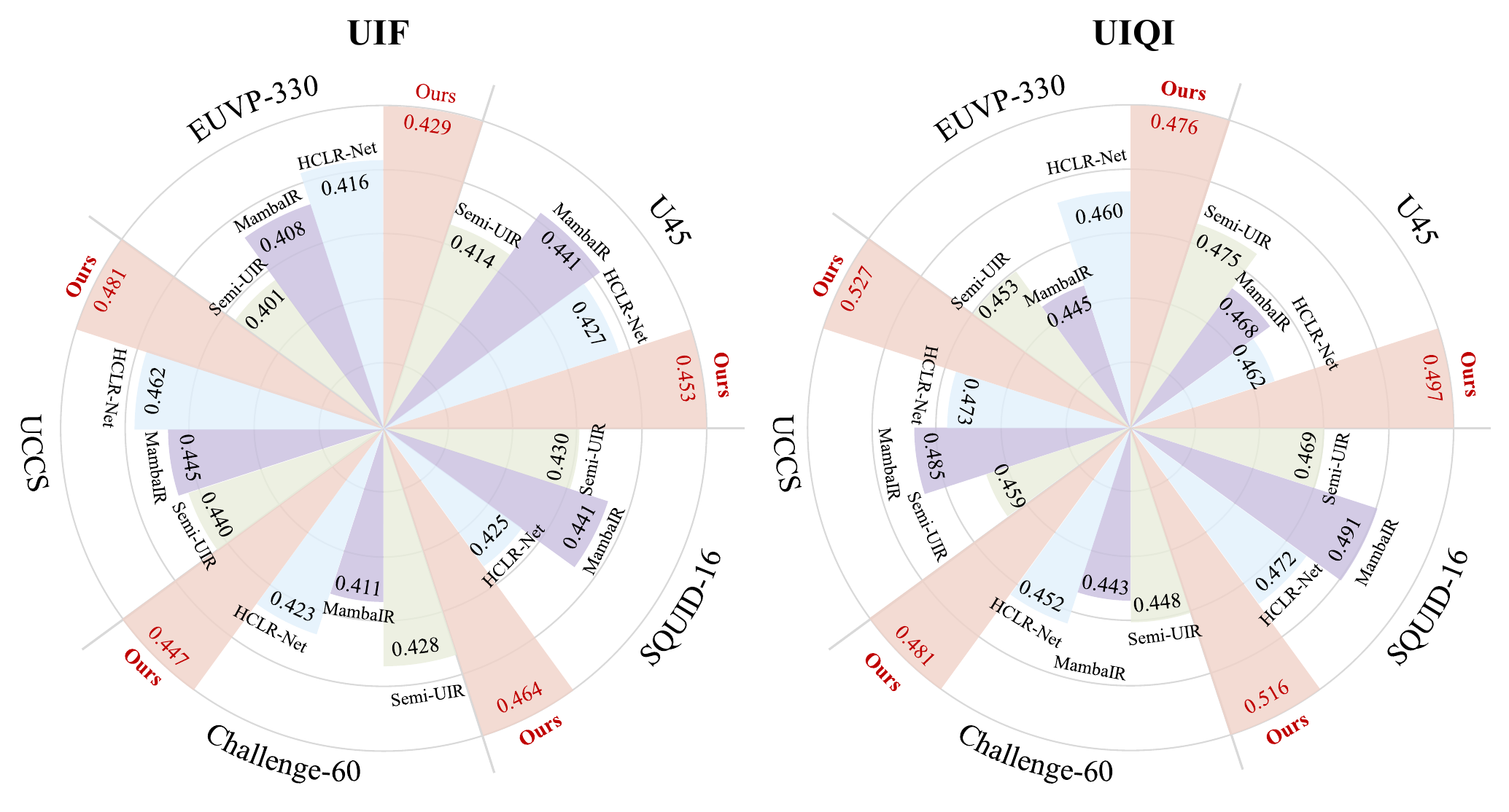}}
  \captionsetup{skip=0pt}
	\caption{Comparison of UIF \cite{UIF} and UIQI \cite{UIQI} scores for state-of-the-art methods on non-reference underwater image datasets.}
	\label{fig:UIF}
	\vspace{-1em}
\end{figure}

\begin{figure*}[!htp]  
	\centerline{\includegraphics[page=1,trim = 0mm 0mm 0mm 0mm, clip, width=1\linewidth]{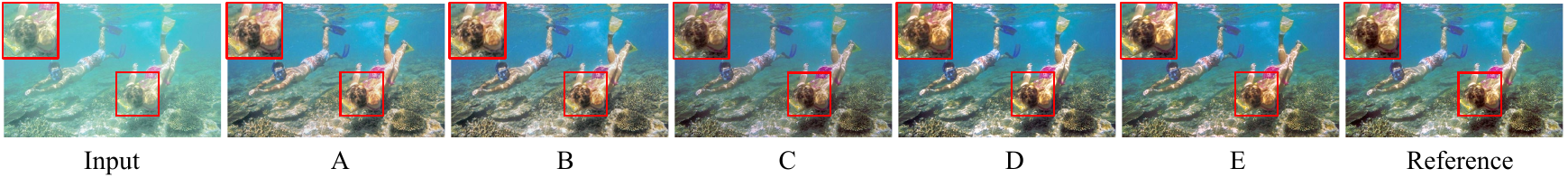}}
  \captionsetup{skip=0pt}
	\caption{Visual results comparison of various network settings. Index A serves as the baseline. Index B extends Index A by incorporating Depth Anything V2. Index C builds upon Index B by integrating the Mamba Mixture-of-Experts Block (MMoEB). Index D further enhances Index C with the Spatial-Frequency Prior Generator (SFPG). Finally, Index E, denoted as UniUIR, represents the full proposed model, augmenting Index D with a Latent Diffusion Model (LCDM).}  
	\label{fig:ablation1}
	\vspace{-1em}
\end{figure*}

\subsection{Qualitative Comparison}
\label{sect:Multi-task_Results}
As shown in Tab.~\ref{tab:ref}, early deep UIR methods like FUnIE \cite{FUnIE} and S-UWnet \cite{S-UWNet} with shallow architectures yield suboptimal results. While simpler methods such as WaterNet \cite{UIEB} benefit from preprocessed inputs (e.g., white balance, gamma correction), more advanced architectures (PUGAN \cite{PUGAN}, PUIE-Net \cite{PUIE-Net}, Ushape \cite{Ushape}) and learning strategies (NU2Net \cite{NU2Net} with URanker-based loss) achieve better performance. Our UniUIR demonstrates superior performance compared to recent SOTA methods, including Semi-UIR \cite{Semi-UIR}, HCLR-Net \cite{zhou_IJCV_HCLR}, DiffUIR\cite{DiffUIR}, and AdaIR \cite{AdaIR}, across six metrics on two datasets.

To assess robustness, we also evaluated performance on non-reference datasets (Tab. \ref{tab:noref}). Specifically, all methods are trained on the UIEB-T dataset and then directly tested on the non-reference datasets. In addition to the commonly used non-reference underwater image quality metrics UIQM and UCIQE, we also incorporate the URanker metric. While higher UIQM and UCIQE scores generally suggest better contrast and color vibrancy, these metrics can be biased, potentially overlooking color shifts and artifacts, as highlighted in \cite{NU2Net}. The URanker score provides a more comprehensive evaluation of underwater image quality. Although MLLE \cite{MLLE}, NU2Net\cite{NU2Net}, and  HCLR-Net \cite{zhou_IJCV_HCLR} tend to achieve higher UCIQE and UIQM scores, Fig. \ref{fig:T90} reveals limitations: MLLE introduces chromatic aberrations, while NU2Net and HCLR-Net struggle with accurate color rendition, particularly noticeable in the Challenge-60 \cite{UIEB} and EUVP-330\cite{EUVP} examples. Both Semi-UIR \cite{Semi-UIR} and our proposed UniUIR achieve high UIQM and UCIQE scores, along with strong URanker performance. Importantly, our method explicitly addresses the co-occurrence of multiple distortions in underwater scenes, restoring images in a unified manner. This approach enhances overall perceptual quality while preserving local details, resulting in improved contrast and vibrant colors, ultimately leading to superior performance on these challenging non-reference datasets. 
Fig. \ref{fig:UIF} demonstrates that our method achieves superior UIF and UIQI scores compared to competing approaches across five benchmark datasets (U45, EUVP-330, SQUID-16, Challenge-60, and UCCS). This consistent performance advantage across diverse no-reference evaluation scenarios confirms the robustness and effectiveness of our approach in enhancing underwater imagery without requiring reference images.

\begin{table*}[!tbp] 
\setlength{\abovecaptionskip}{2pt}
  \caption{
  Effectiveness of components on the T90 \cite{UIEB} dataset. DAM V2 denotes Depth Anything Model V2, MMoEB denotes Mamba Mixture-of-Experts Block, SFPG denotes Spatial-Frequency Prior Generator, and LCDM denotes Latent Conditional Diffusion Model.
  }

  \label{tab:components}
  
  \centering
  \renewcommand\arraystretch{1.2}	
    {\begin{tabular}{c|cccc|cccccccc} 
  
\toprule[1pt]
\toprule[0.5pt]
  
    Index
    & DAM V2
    & MMoEB
    & SFPG
    & LCDM
    &  PSNR$\uparrow$
    &  SSIM$\uparrow$
    &  LPIPS$\downarrow$
    &  UCIQE$\uparrow$
    &  UIQM$\uparrow$
    &  URanker$\uparrow$
    &  {UIF$\uparrow$}
    &  {UIQI$\uparrow$}
    \\

    \hline
    A (Baseline)
    &\textcolor{red}{\ding{55}}
    &\textcolor{red}{\ding{55}}
    &\textcolor{red}{\ding{55}} 
    &\textcolor{red}{\ding{55}} 
    & 24.37 & 0.928 & 0.131 & 0.581 & 3.059  & 1.894  
    &{0.453}
    &{0.459}
    \\

    \cdashline{1-13}
    B
    &\textcolor{green}{\ding{51}}
    &\textcolor{red}{\ding{55}}
    &\textcolor{red}{\ding{55}}  
    &\textcolor{red}{\ding{55}} 
    & 24.56 & 0.930 & 0.124 & 0.589 & 2.985 & 1.911 
    &{0.467}
    &{0.475}
    \\

    C
    &\textcolor{green}{\ding{51}}
    &\textcolor{green}{\ding{51}}
    &\textcolor{red}{\ding{55}} 
    &\textcolor{red}{\ding{55}} 
    & 24.81 & 0.932 & 0.120 & 0.597 & 3.034 &1.986 
    &{0.466}
    &{0.478}
    \\

    D
    &\textcolor{green}{\ding{51}}
    &\textcolor{green}{\ding{51}}
    &\textcolor{green}{\ding{51}}
    &\textcolor{red}{\ding{55}}
    & 24.97 & 0.932 & 0.118 & \bf0.609 & 3.057 & 2.071
    &{0.472}
    &{0.483}
    \\

    \rowcolor{my_color}
    E
    &\textcolor{green}{\ding{51}}
    &\textcolor{green}{\ding{51}}
    &\textcolor{green}{\ding{51}}  
    &\textcolor{green}{\ding{51}} 
    & \bf25.11 & \bf0.933 & \bf0.112 & 0.601 & \bf3.117 & \bf2.104 
    &{\bf0.481}
    &{\bf0.496}    
    \\

  \bottomrule[1pt]
  \end{tabular}}
  \vspace{-2em}
\end{table*}

\subsection{Quantitative Comparisons} 
This section presents a comprehensive visual comparison across reference and non-reference test sets. Beginning with the T90 dataset \cite{UIEB} (Fig. \ref{fig:T90}), we observe that the traditional method MLLE\cite{MLLE} often introduces chromatic aberrations and struggles with color distortions. Deep learning methods generally outperform MLLE, but limitations remain. For example, the WaterNet\cite{UIEB} struggles with yellow color casts. Ucolor\cite{Ucolor} suffers from a loss of fine texture details. NU2Net\cite{NU2Net} and HCLR-Net\cite{zhou_IJCV_HCLR} occasionally introduce chromatic aberrations, as seen in the image 2 and image 4 samples of Fig. \ref{fig:T90}, indicating difficulties in controlling color. In contrast, our proposed method effectively mitigates these issues, producing enhanced images with minimal color deviations, improved contrast, and enhanced structural details. Furthermore, due to our network's ability to address multiple distortions simultaneously, we achieve superior restoration results in challenging low-light conditions, as exemplified by image 5.

We also conduct visual comparisons across five non-reference underwater datasets. As shown in Fig. \ref{fig:noref}, many traditional and deep learning methods struggle with these datasets due to the diverse and complex distortions present. For example, MLLE frequently introduces color distortions. WaterNet and Semi-UIR struggle with haze, as seen in images 8 and 16. Additionally, WaterNet performs poorly in low-light conditions, such as image 11. NU2Net exhibits minor color casts on the Challenge-60\cite{UIEB}, UCCS\cite{UCCS}, and EUVP-330 datasets\cite{EUVP}. Ucolor and HCLR-Net struggle with yellow color casts. Furthermore, HCLR-Net performs poorly on blurry and distorted images from the U45 dataset, often lacking fine details. In contrast, our proposed UniUIR achieves the best overall visual results.

\begin{table}[!htbp] 
\setlength{\abovecaptionskip}{2pt}
  \caption{Effect of number of experts on the T90 \cite{UIEB} dataset.}
  \label{tab:experts}
  \centering
  \renewcommand\arraystretch{1.2}	
  \tabcolsep=0.15cm 
    \resizebox{\linewidth}{!}{\begin{tabular}{c|cccccc} 
  
\toprule[1pt]
\toprule[0.5pt]
  
    Number
    &  PSNR$\uparrow$
    &  SSIM$\uparrow$
    &  LPIPS$\downarrow$
    &  UCIQE$\uparrow$
    &  UIQM$\uparrow$
    &  URanker$\uparrow$
    \\

    \hline
    $n$=1
    & 24.71 & 0.931 & 0.121 & 0.597 & 3.064  & 1.938  
    \\

    $n$=2
    & 24.94 & 0.932 & 0.117 & 0.602 & 3.025 & 1.995 
    \\

    \rowcolor{my_color}
    $n$=3
    & \bf25.11 & \bf0.933 & \bf0.112 & 0.601 & \bf3.117 & \bf2.104 
    \\

    $n$=4
    & 25.01 & 0.932 & 0.115 & \bf0.607 & 2.993 &2.052
    \\

  \bottomrule[1pt]
  \end{tabular}}
  \vspace{-1.5em}
\end{table}

\begin{table}[tbp]
\setlength{\abovecaptionskip}{2pt}
  \caption{Effect of the number of MMoEBs on the T90 \cite{UIEB} dataset.}
  
  \label{tab:MMoEBs}
  
  \centering
  \renewcommand\arraystretch{1.3}	
  \tabcolsep=0.15cm 
    \resizebox{\linewidth}{!}{\begin{tabular}{c|cc|cccc} 
  
\toprule[1pt]
\toprule[0.5pt]
  
    Setting
    &  Param. (M)$\downarrow$
    &  FLOPs (G)$\downarrow$
    &  UCIQE$\uparrow$
    &  UIQM$\uparrow$
    &  UIF$\uparrow$
    &  UIQI$\uparrow$
    \\

    \hline
    $[2,4,4,6]$
    & 23.78 & 63.41  & 0.591 & 3.037  & 0.476  & 0.485
    \\

\rowcolor{my_color}
    $[3,5,6,6]$ 
    & 29.19 & 78.07  & 0.601 & \bf3.117 & 0.481 &  0.496
    \\

    $[4,6,6,8]$
    & 34.58 & 91.26  & \bf0.604 & 3.088 & \bf0.487 & \bf0.504
    \\

  \bottomrule[1pt]
  \end{tabular}}
  \vspace{-1.5em}
\end{table}

\begin{table}[!htbp]
\setlength{\abovecaptionskip}{2pt}
  \caption{Effect of loss function ablation on the T90 \cite{UIEB} dataset.}
  \label{tab:loss}
  \centering
  \renewcommand\arraystretch{1.2}	
  \tabcolsep=0.15cm 
    \resizebox{\linewidth}{!}{\begin{tabular}{l|cccccc} 
  
\toprule[1pt]
\toprule[0.5pt]
  
    Setting
    &  PSNR$\uparrow$
    &  SSIM$\uparrow$
    &  UCIQE$\uparrow$
    &  UIQM$\uparrow$
    &  UIF$\uparrow$
    &  UIQI$\uparrow$
    \\

    \hline
    w/o $\mathcal{L}_{depth}$
    & 24.72 & 0.930  & 0.592 & 2.981  & 0.465  & 0.477
    \\

    w/o $\mathcal{L}_{grad}$
    & 24.93 & 0.932  & \bf0.603 & 3.069 & 0.472 & 0.481
    \\

    w/o $\mathcal{L}_{diff}$
    & 24.76 & 0.929  & 0.587 & 3.016 & 0.467 & 0.484
    \\

    \rowcolor{my_color}
    All loss
    & \bf25.11 & \bf0.933  & 0.601 & \bf3.117 & \bf 0.481 &  \bf0.496
    \\

  \bottomrule[1pt]
  
  \end{tabular}}
  \vspace{-1.5em}
\end{table}

\subsection{Ablation Study} %
We conduct several ablation experiments to demonstrate the effectiveness of each component in the proposed UniUIR. All ablation experiments are trained on the UIEB-T \cite{UIEB} dataset and tested on T90 \cite{UIEB} dataset.

\subsubsection{Effects of Different Components} 
As shown in Tab. \ref{tab:components}, we evaluate the effectiveness of various components through a comparison with baseline (Index A) that excludes our modules. Integrating depth information priors derived from Depth Anything V2 demonstrably improves restoration quality, increasing PSNR, SSIM, and reducing LPIPS by 0.19 dB, 0.002, and 0.004, respectively. No-reference metrics UCIQE and URanker also show improvement. The introduction of MMoEB enhances the model's ability to handle multiple distortions simultaneously, leading to substantial gains across both full-reference and no-reference metrics. Further incorporating SFPG and LCDM yields additional performance benefits, indicating that spatial-frequency degradation priors and latent diffusion modeling contribute to more effective underwater image restoration. Notably, UCIQE and UIQM trends don't always align with other metrics (e.g., UIQM in Indexs A and B, and UCIQE in Indexs D and E), potentially echoing findings in \cite{NU2Net} regarding limitations of UCIQE and UIQM in reflecting perceptual underwater image quality. Finally, the visual results in Fig. \ref{fig:ablation1} demonstrate that our proposed approach (Index E) achieves the best visual quality.

\subsubsection{Effects of Number of Experts} 
Tab. \ref{tab:experts} presents the effectiveness of varying the number of experts on the T90 dataset. It can be observed that the trends for PSNR, SSIM, LPIPS, and URanker are consistent, all achieving their best values when $n=3$. Although the trends for UCIQE and UIQM differ slightly from the aforementioned metrics, the best UIQM value is also obtained at $n=3$. These findings suggest that selecting $n=3$ experts strikes the optimal balance across all metrics, making it the most suitable configuration for the experiments.
\subsubsection{Effects of Number of MMoEBs} 
As shown in Tab. \ref{tab:MMoEBs}, the [3,5,6,6] configuration achieves the best trade-off between performance and efficiency. It yields the highest UIQM score while maintaining moderate model complexity. Although the [4,6,6,8] variant achieves slightly better results in UCIQE, UIF, and UIQI, it incurs significantly higher computational cost. These results suggest that increasing the number of experts does not guarantee consistent performance gains and must be carefully balanced with efficiency constraints.
\subsubsection{Effects of Each Loss Term} 
As shown in Table~\ref{tab:loss}, removing any loss leads to performance drops in most metrics, particularly in PSNR and SSIM, indicating their overall effectiveness. Notably, the model without $\mathcal{L}_{grad}$
achieves a slightly higher UCIQE (0.603 vs. 0.601), suggesting that gradient supervision may slightly over-sharpen structures under this metric. However, the full model achieves the best balance across all metrics, demonstrating that the combined use of $\mathcal{L}_{depth}$, $\mathcal{L}_{grad}$, and $\mathcal{L}_{diff}$ yields the most robust and perceptually favorable restoration.

\begin{figure}[!htp]  
	\centerline{\includegraphics[page=1,trim = 0mm 0mm 0mm 0mm, clip, width=1\linewidth]{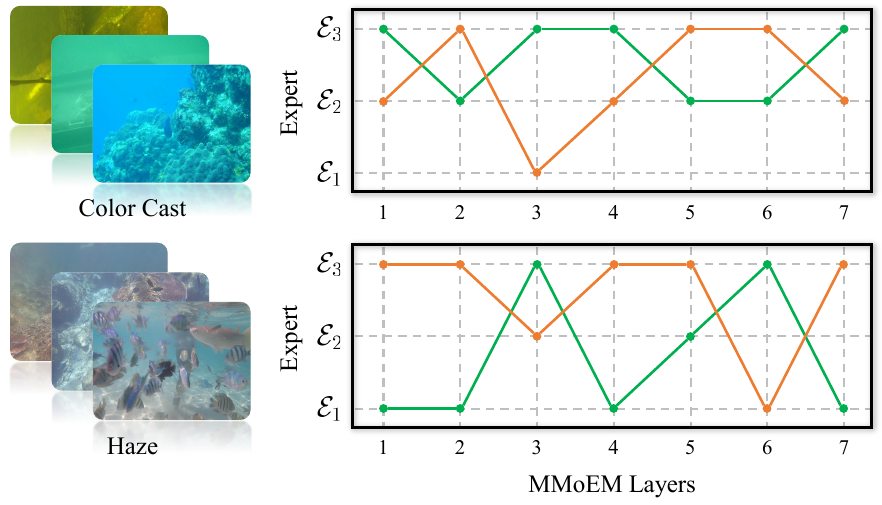}}
  \captionsetup{skip=0pt}
	\caption{Expert selection in MMoEM under various degradation types.}  %
	\label{fig:moe}
	\vspace{-1em}
\end{figure}

\begin{figure*}[!htp]  
	\centerline{\includegraphics[page=1,trim = 0mm 0mm 0mm 0mm, clip, width=1\linewidth]{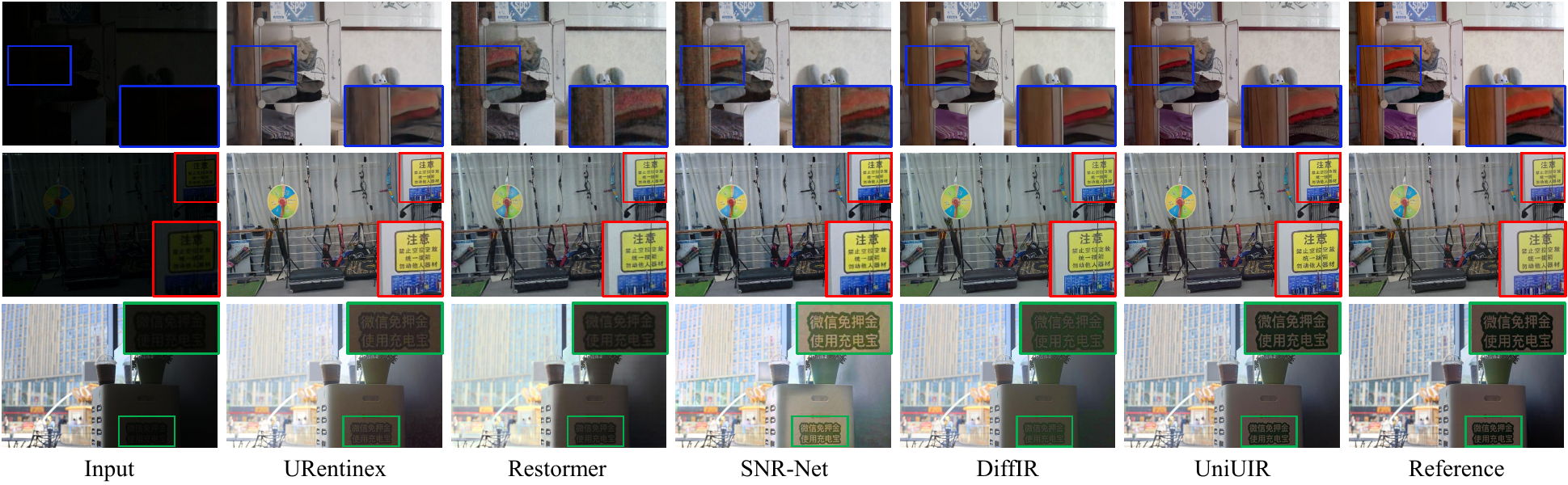}}
  \captionsetup{skip=0pt}
	\caption{Qualitative results on LOL-v1 (top), LOL-v2-real (middle), and BAID (bottom).}  
	\label{fig:LLIE}
	\vspace{-1em}
\end{figure*}

\begin{figure*}[!htp]  
	\centerline{\includegraphics[page=1,trim = 0mm 0mm 0mm 0mm, clip, width=1\linewidth]{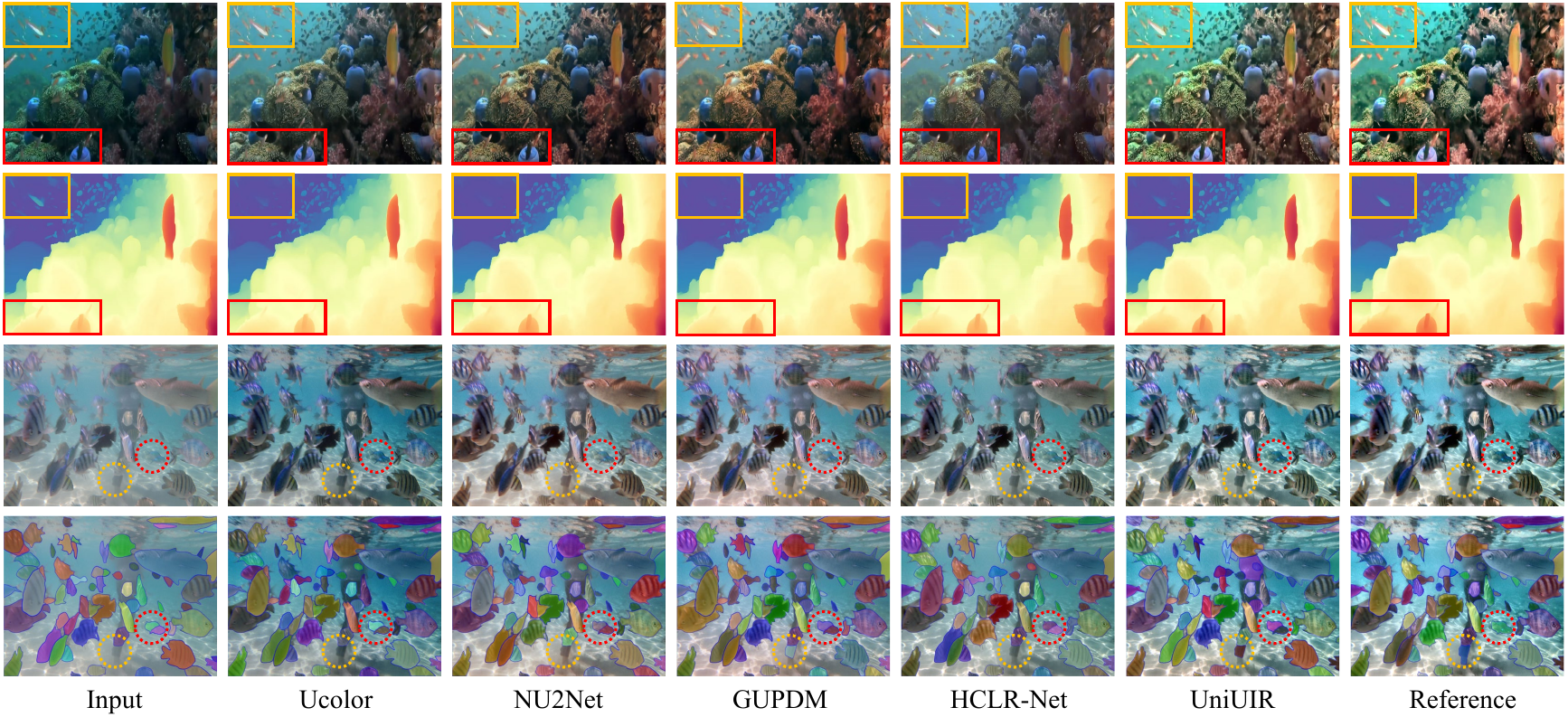}}
  \captionsetup{skip=0pt}
	\caption{The depth maps and semantic segmentation results generated by Depth Anything V2 \cite{Depth_V2} and the Segment Anything Model \cite{SAM} are presented in Rows 2 and 4, respectively. These images are derived from the original underwater images input and those restored using Ucolor, NU2Net, GUPDM, HCLR-Net, the proposed UniUIR, and corresponding reference images. For ease of comparison, the restored images are also displayed in Rows 1 and 3.}  %
	\label{fig:sam}
	\vspace{-1em}
\end{figure*}

\begin{figure*}[!htp]  
	\centerline{\includegraphics[page=1,trim = 0mm 0mm 0mm 0mm, clip, width=1\linewidth]{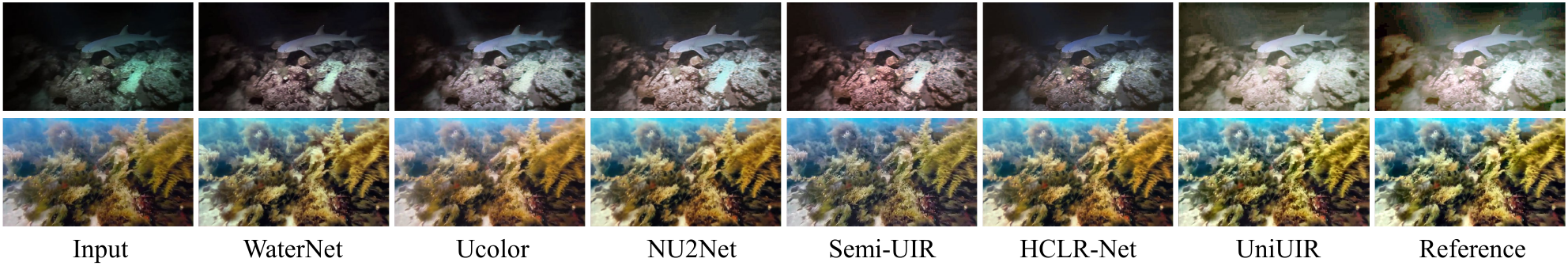}}
  \captionsetup{skip=0pt}
	\caption{Demonstration of failure cases for the proposed UniUIR and state-of-the-art underwater image restoration methods on the T90 dataset.}  %
	\label{fig:failcase}
	\vspace{-1.5em}
\end{figure*}

\begin{table}[!htbp]
\setlength{\abovecaptionskip}{2pt}
  \caption{Comparison of model complexity and time efficiency across different methods in terms of Parameters (M), FLOPs (G), and Inference Time (ms).}
  \label{tab:complexity}
  \centering
  \renewcommand\arraystretch{1.2}	
  \tabcolsep=0.15cm 
    \resizebox{0.85\linewidth}{!}{\begin{tabular}{l|cccccc} 
  
\toprule[1pt]
\toprule[0.5pt]
  
    Method
    &  Param. (M)$\downarrow$
    &  FLOPs (G)$\downarrow$
    &  Time (ms)$\downarrow$
    \\

    \hline

    WaterNet \cite{UIEB}
    & 24.81 
    & 193.7 
    & 314  
    \\

    FUnIE \cite{FUnIE}
    & 7.7 
    & 10.7 
    & 38  
    \\

    Ucolor \cite{Ucolor}
    & 157.4
    & 34.68 
    & 142 
    \\

    PUIE-Net \cite{PUIE-Net}
    & 10.69 
    & 30.09 
    & 262  
    \\

    U-shape \cite{Ushape}
    & 65.60 
    & 66.20 
    & 97 
    \\

    PUGAN \cite{Semi-UIR}
    & 95.66 
    & 72.05 
    & 276
    \\

    NU2Net \cite{NU2Net}
    & 3.1
    & 10.4
    & 67
    \\

    Semi-UIR \cite{Semi-UIR}
    & 12.78 
    & 36.46 
    & 147
    \\

    PromptIR \cite{PromptIR}
    & 32.96 
    & 158.14 
    & 111
    \\

    HCLR-Net \cite{zhou_IJCV_HCLR}
    & 4.87 
    & 401.97 
    & 239
    \\

    MambaIR \cite{MambaIR}
    & 26.78 
    & 67.38 
    & 133
    \\

    AdaIR \cite{AdaIR} 
    & 28.77 
    & 147.5 
    & 116
    \\

    \rowcolor{my_color}
    UniUIR (Ours) 
    & 29.19 
    & 78.07 
    & 105
    \\

  \bottomrule[1pt]
  \end{tabular}}
  \vspace{-1.5em}
\end{table}

\begin{table*}[!htbp]
\setlength{\abovecaptionskip}{2pt}
  \caption{Comparison of performance, model complexity, and efficiency across UniUIR components.}
  
  \label{tab:complexity_each}
  
  \centering
  \renewcommand\arraystretch{1.2}	
  \tabcolsep=0.1cm 
    \resizebox{0.7\linewidth}{!}{\begin{tabular}{l|ccc|ccc} 
  
\toprule[1pt]
\toprule[0.5pt]
  
    Method
    &  Param. (M)$\downarrow$
    &  FLOPs (G)$\downarrow$
    &  Time (ms)$\downarrow$
    & URanker$\uparrow$
    & UIF$\uparrow$
    & UIQI$\uparrow$
    \\

    \hline

    UniUIR$_{S1}$  
    & 27.95
    & 70.12 
    & 84  
    & 2.167 
    & 0.493 
    & 0.511 
    \\

    \cdashline{1-7}

    UniUIR$_{S2}$-V1 (w/o DAM V2)
    & 29.19   
    & 65.56
    & 80
    & 1.996
    & 0.483
    & 0.492 
    \\

    UniUIR$_{S2}$-V2 (w/o SFPG)
    & 26.41 
    & 63.24 
    & 93
    & 2.019 
    & 0.474 
    & 0.487 
    \\

    UniUIR$_{S2}$-V3 (w/o LCDM)
    & 27.56 
    & 68.33 
    & 75
    & 2.071 
    & 0.472 
    & 0.483 
    \\

    UniUIR$_{S2}$-V4 (w/ MMoE-UIR)
    & 23.37 
    & 51.32 
    & 61
    & 1.924 
    & 0.465 
    & 0.478 
    \\

    \rowcolor{my_color}
    UniUIR$_{S2}$ (Ours)
    & 29.19 
    & 78.07 
    & 105
    & 2.104 
    & 0.481
    & 0.496 
    \\

  \bottomrule[1pt]
  \end{tabular}}
  \vspace{-1.5em}
\end{table*}

\begin{table*}[!htbp] 
\setlength{\abovecaptionskip}{2pt}
  \caption{Comparison of performance and inference time across different settings.}
  \label{tab:complexity_depthpro}
  \centering
  \renewcommand\arraystretch{1.2}	
    \resizebox{0.85\linewidth}{!}{\begin{tabular}{l|ccccccc|cc} 
  
\toprule[1pt]
\toprule[0.5pt]
  
    Setting
    & PSNR$\uparrow$
    & SSIM$\uparrow$
    & UCIQE$\uparrow$
    & UIQM$\uparrow$
    & URanker$\uparrow$
    & UIF$\uparrow$
    & UIQI$\uparrow$
    &  FLOPs (G)$\downarrow$
    &  Time (ms)$\downarrow$
    \\

    \hline

    \rowcolor{my_color}\bf Baseline
    & 25.11
    & 0.933
    & 0.601
    & 3.117
    & 2.104
    & 0.481
    & 0.496
    & 78.07
    & 105 
    \\

    DAM V2$\rightarrow$Depth-Pro
    & 25.23 
    & 0.933
    & 0.605
    & 3.098
    & 2.134
    & \bf0.490
    & 0.501
    & 64.35
    & 91  
    \\

    4-Step$\rightarrow$2-Step 
    & 24.75 
    & 0.931
    & 0.585
    & 2.994
    & 1.932
    & 0.465
    & 0.479
    & \bf51.24
    & \bf73  
    \\

    4-Step$\rightarrow$8-Step
    & \bf25.34 
    & \bf0.934
    & \bf0.609
    & \bf3.145
    & \bf2.138
    & 0.486
    & \bf0.504
    & 112.37
    & 154
    \\

  \bottomrule[1pt]
  \end{tabular}}
  \vspace{-1em}
\end{table*}

\begin{table*}[tbp] 
\setlength{\abovecaptionskip}{2pt}
  \caption{
  Performance comparisons on low-light and backlit image enhancement tasks.}

  \label{tab:LLIE}
  
  \centering
    \renewcommand\arraystretch{1.1}	
    {\begin{tabular}{l|ccc|ccc|ccc|ccc} 
  
\toprule[1pt]
\toprule[0.5pt]
  
    \multirow{2}{*}{Method}
    & \multicolumn{3}{c|}{\bf LOL-v1 \cite{LOL}} 
    & \multicolumn{3}{c|}{\bf LOL-v2-real \cite{LOL_v2}}
    & \multicolumn{3}{c|}{\bf SID \cite{SID}}
    & \multicolumn{3}{c}{\bf BAID \cite{BAID}}
    \\ \cline{2-13} 

    {}
    & PSNR$\uparrow$ & SSIM$\uparrow$ & FID$\downarrow$ 
    & PSNR$\uparrow$ & SSIM$\uparrow$ & FID$\downarrow$
    & PSNR$\uparrow$ & SSIM$\uparrow$ & FID$\downarrow$
    & PSNR$\uparrow$ & SSIM$\uparrow$ & FID$\downarrow$
    \\
    
    \hline
    %
    URetinex \cite{URetinex} 
    & 21.33 & 0.835  & 85.59
    & 20.44 & 0.806  & 76.74
    & 22.09 & 0.633  & 71.58
    & 19.08 &0.845   & 42.26\\

    Restormer \cite{Restormer} 
    & 22.43 &0.823  & 78.75
    & 19.94 &0.827  & 114.35
    & 22.27 &0.649  & 75.47
    & 21.07 &0.832  & 41.17\\

    SNR-Net \cite{SNR-Net} 
    &24.61  &0.842  & 66.47
    &21.48  &0.849  & 68.56
    &22.87  &0.625  & 74.78
    &20.86  &0.860  & 39.73\\

    DiffIR \cite{DiffIR} 
    &23.15  &0.828  & 70.13
    &21.15  &0.816  & 72.33
    &23.17  &0.640  & 78.80
    &21.10  &0.835  & 40.35\\

    Diff-Retinex \cite{Diff-Retinex} 
    & 21.98  & \bf{{0.852}}  & \bf51.33
    & 20.17  & 0.826  & 46.67
    & 23.62  & 0.665  & \bf58.93
    & 22.07  & 0.861  & 38.07\\

    \rowcolor{my_color}\bf UniUIR 
    & \bf{24.83} & {0.851}  & {55.19}
    & \bf{22.69} & \bf{0.850}  & \bf 46.23
    & \bf 24.67  & \bf 0.685   &  60.64
    & \bf 22.21  & \bf 0.864   & \bf36.23\\

  \bottomrule[1pt]
  \end{tabular}}
  \vspace{-1em}
\end{table*}

\subsection{Discussion} 
\subsubsection{Performance vs. Efficiency}
As summarized in Tab.~\ref{tab:complexity}, UniUIR achieves a favorable trade-off between restoration quality and computational efficiency. It is more efficient than WaterNet, Ucolor, and several GAN-based methods in terms of FLOPs and latency, while outperforming them in visual quality. Although NU2Net runs faster, its significantly lower capacity may limit restoration performance. Compared to recent approaches like PromptIR and AdaIR, UniUIR offers better efficiency and accuracy.

As shown in Tab.~\ref{tab:complexity_each}, 
the full UniUIR model achieves the best restoration quality at the cost of higher computational load. 
Notably, UniUIR$_{S1}$ outperforms UniUIR$_{S2}$ due to its use of reference image information during inference, which is not available in the latter. The removal of key components, including DAM V2, SFPG, or LCDM, or their substitution with lighter alternatives such as MMoE-UIR, reduces complexity and latency but leads to performance degradation. This result confirms their essential role in maintaining high-quality restoration.

As shown in Tab. \ref{tab:complexity_depthpro}, the results indicate that while increasing the number of diffusion steps can further enhance the restoration quality across various metrics, it also significantly increases the inference time. For instance, the 4-Step→8-Step setting achieves the highest PSNR but requires 154 ms for inference, which is considerably longer than the baseline.
Conversely, reducing the number of diffusion steps can drastically reduce the inference time. However, this comes at the cost of slightly degraded performance, with PSNR dropping to 24.75 dB.
Interestingly, the DAM V2→Depth-Pro \cite{Depth-pro} setting offers a balance between performance and efficiency. It not only improves all image quality metrics compared to the baseline but also reduces the inference time. This suggests that integrating more advanced and lightweight depth estimation models into the architecture can effectively improve both restoration quality and computational efficiency.
\subsubsection{Expert Selection Analysis}
As shown in Fig. \ref{fig:moe}, $\mathcal{E}_{2}$ and $\mathcal{E}_{3}$ are more readily activated when the primary degradation is color cast,
whereas $\mathcal{E}_{1}$ and $\mathcal{E}_{3}$ are more frequently activated under haze degradation.
This demonstrates that the model can adaptively and dynamically route different degradation patterns to distinct expert combinations, effectively implementing a decomposition-and-specialization strategy within the network. This adaptive expert activation mechanism ensures that the model does not merely memorize a single restoration strategy but instead utilizes specialized sub-networks to handle specific distortions, thereby enhancing both the interpretability and effectiveness of the restoration process.
\subsubsection{Application on Other Low-level Vision Tasks}
To evaluate UniUIR on low-light and backlit enhancement, we retrained it and compared with state-of-the-art methods on LOL-v1~\cite{LOL}, LOL-v2-real~\cite{LOL_v2}, SID~\cite{SID}, and BAID~\cite{BAID}. 
As shown in Tab.~\ref{tab:LLIE}, UniUIR achieves superior performance in PSNR, SSIM, and perceptual quality, demonstrating its robustness and generalizability. 
Qualitative results in Fig.~\ref{fig:LLIE} show that UniUIR effectively corrects illumination and enhances texture, even under complex lighting or high-frequency details. 
Unlike existing methods, UniUIR leverages adaptive feature extraction and spatial-frequency priors to handle challenging conditions, delivering consistently high-quality enhancement.
\subsubsection{Application on High-level Vision Tasks}
We evaluate the impact of restored underwater images on downstream tasks: depth estimation and image segmentation. 
\textbf{i)} For depth estimation, we use Depth Anything V2~\cite{Depth_V2} to extract depth maps from restored images. As shown in Fig.~\ref{fig:sam} (second row, orange/red boxes), UniUIR produces results closest to the reference, demonstrating superior depth consistency and distortion correction. 
\textbf{ii)} For segmentation, we used restored images from the T90 dataset produced by various UIR methods and directly applied the Segment Anything Model (SAM) \cite{SAM}. As shown in the fourth row (yellow dashed box), both HCLR-Net~\cite{zhou_IJCV_HCLR} and UniUIR correctly segment the occluded diver's fins, while UniUIR yields superior visual quality. Notably, in the red dashed box, the reference image mismerges two fish, while all UIR methods (even the input) correctly separate them, highlighting the need to balance perceptual quality and machine vision performance.

These results show that UniUIR enhances not only visual quality but also the utility of restored images for downstream tasks, advocating for UIR methods that optimize both human and machine perception.

\subsection{Limitation and Future Work}
Although our proposed method consistently outperforms state-of-the-art approaches in underwater image enhancement, certain limitations persist. As illustrated in Fig. \ref{fig:failcase}, existing methods struggle with enhancing underwater images under extreme low-light and blurry conditions. 
This difficulty can be attributed to the fact that most current datasets emphasize common distortions such as color shifts and low contrast, with limited examples of extreme scenarios like those encountered in low-light environments. Consequently, the learning capacity of models is constrained, leading to challenges in handling more complex conditions. The scarcity of such samples not only limits the model’s exposure to severe degradation patterns but also biases the training objective toward average-case performance, reducing its robustness in rare yet critical situations.
Moreover, our method leverages the pre-trained Depth Anything V2 model, which has been primarily trained on natural scenes containing minimal underwater data. Directly applying its pre-trained weights may result in suboptimal capture of depth features specific to underwater settings.

To address these issues, future work could employ image generation techniques to augment the training dataset with synthetic extreme underwater scenes. This would enrich the training set with diverse and realistic samples, significantly enhancing the model's generalization ability in complex scenarios. Furthermore, exploring efficient fine-tuning strategies for depth extraction tailored to underwater environments can be highly beneficial. For example, employing Low-Rank Adaptation (LoRA) techniques to fine-tune pre-trained depth extraction models \cite{LoRA_Depth} would enable the recovery network to better accommodate underwater-specific characteristics such as light variations, turbidity, and depth inconsistencies. This approach would enhance the model's robustness and performance under challenging underwater conditions.
  

\section{Conclusion}
In this paper, we systematically review the current challenges in underwater image restoration and propose UniUIR, an all-in-one approach designed to address the complex underwater scenario. To effectively harness the interdependencies among various degradations in underwater environments, we developed the Mamba Mixture-of-Experts module. This module allows each expert to specialize in distinct aspects of degradation and collaboratively infer task-specific priors while preserving both local and global feature representations. To enhance the model's ability to identify inconsistencies between distortions across different image regions, we integrate scene depth information using the Depth Anything V2 model. Additionally, we propose a spatial-frequency prior generator that extracts degradation priors from both spatial and frequency domains via task-adaptive routing. This mechanism enables conflicting tasks to utilize distinct network paths, thereby reducing task interference. Compared to state-of-the-art underwater image restoration methods, UniUIR delivers superior results, as evidenced by extensive qualitative and quantitative evaluations.


%
%

\bibliographystyle{IEEEtran}
\bibliography{IEEEabrv,UniUIR}

\begin{IEEEbiography}[{\includegraphics[width=1in,height=1.25in,clip,keepaspectratio]{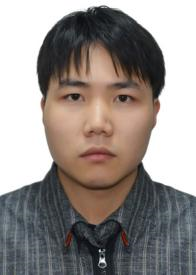}}]{Xu Zhang}
received the B.E. degree from Jianghan University, Wuhan, China, in 2020.  He obtained the M.E. degree from Guangdong University of Technology, Guangzhou, China, in 2024. Currently, he is pursuing Ph. D. degree in Wuhan University, Wuhan, China. 

His research interests include image restoration and image/video quality assessment.
\end{IEEEbiography}

\begin{IEEEbiography}[{\includegraphics[width=1in,height=1.25in,clip,keepaspectratio]{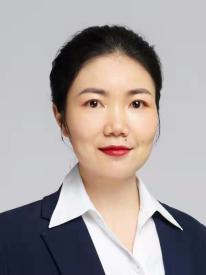}}]{Huan Zhang} received the B.S. degree from Civil Aviation University of China, Tianjin, China, in 2010, M.S. degree from Tsinghua University, Beijing, China in 2013, and Ph.D. degree from University of Chinese Academy of Sciences in 2021. She is currently with the School of Information Engineering, Guangdong University of Technology, Guangzhou, China.  

Her research interests include image restoration, 3D image/video quality assessment, and learned image compression.
\end{IEEEbiography}

\begin{IEEEbiography}[{\includegraphics[width=1in,height=1.25in,clip,keepaspectratio]{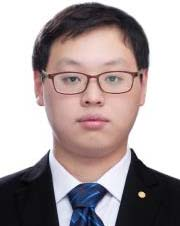}}]{Guoli Wang} received the B.S. degree in electronic information science and technology from the China University of Mining and Technology, Xuzhou, China, in 2011, and the Ph.D. degree in pattern recognition and intelligent systems from the National Laboratory of Pattern Recognition, Institute of Automation, Chinese Academy of Sciences, Beijing, China, in 2017. In 2021, he completed his Post-Doctoral fellowship with the Department of Automation, Tsinghua University, Beijing, China.  

His research interests include computer vision, pattern recognition, and machine learning. 
\end{IEEEbiography}

\begin{IEEEbiography}[{\includegraphics[width=1in,height=1.25in,clip,keepaspectratio]{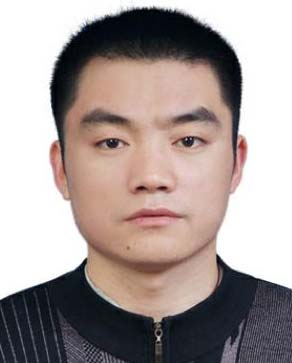}}]{Qian Zhang} received the B.E. and M.S. degrees
from Central South University, Changsha, China, in 2008 and 2011, respectively, and the Ph.D. degree in pattern recognition and intelligent systems from
the Institute of Automation, Chinese Academy of Sciences, Beijing, China, in 2014. 

His research interests include computer vision and machine learning.

\end{IEEEbiography}

\begin{IEEEbiography}
[{\includegraphics[width=1in,height=1.25in,clip,keepaspectratio]{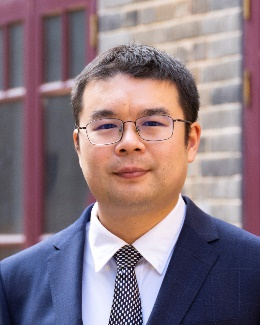}}]{Lefei Zhang}
(Senior Member, IEEE) received the B.S. and Ph.D. degrees from Wuhan University, Wuhan, China, in 2008 and 2013, respectively. He was a Big Data Institute Visitor with the Department of Statistical Science, University College London, U.K., and a Hong Kong Scholar with the Department of Computing, The Hong Kong Polytechnic University, Hong Kong, China. He is a professor with the School of Computer Science, Wuhan University, Wuhan, China, and also with the Hubei Luojia Laboratory, Wuhan, China. His research interests include pattern recognition, image processing, and remote sensing. 

Dr. Zhang serves as an associate editor of IEEE Transactions on Geoscience and Remote Sensing, an associate editor of Pattern Recognition, and an editorial board member of Information Fusion.

\end{IEEEbiography}

\begin{IEEEbiography}[{\includegraphics[width=1in,height=1.25in,clip,keepaspectratio]{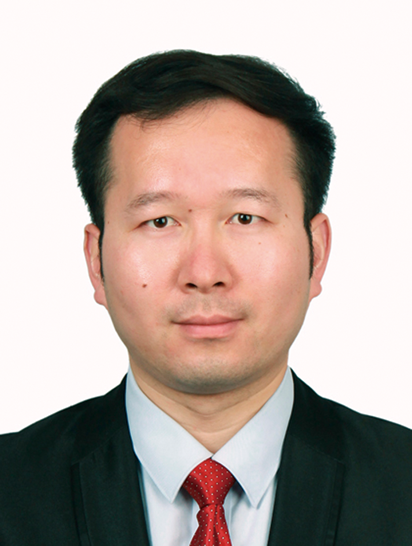}}]{Bo Du} (Senior Member, IEEE) received the Ph.D.
degree in photogrammetry and remote sensing from
the State Key Laboratory of Information Engineering
in Surveying, Mapping and Remote Sensing, Wuhan
University, Wuhan, China, in 2010.
He is currently a Professor with the School of
Computer Science and the Institute of Artificial
Intelligence, Wuhan University, where he is also the
Director of the National Engineering Research Center for Multimedia and Software. He has more than 160 research papers published in the IEEE TPAMI, IEEE TIP, IEEE TCYB, and IEEE TGRS. His major research interests include artificial intelligence, computer vision, large models, and medical image processing.

Dr. Du serves as a Senior Program Committee (PC) Member of the International Joint Conferences on Artificial Intelligence (IJCAI) and the Association
for the Advancement of Artificial Intelligence (AAAI). He received the Highly
Cited Researcher by the Web of Science Group in 2019 and 2020, the IEEE
Geoscience and Remote Sensing Society (GRSS) in 2020 Transactions Prize
Paper Award, the IJCAI Distinguished Paper Prize, the IEEE Data Fusion
Contest Champion, and the IEEE Workshop on Hyperspectral Image and
Signal Processing Best Paper Award in 2018. He has served as the Area Chair
for the International Conference on Pattern Recognition (ICPR). He also serves as an Associate Editor for IEEE TIP, IEEE TNNLS, IEEE TCSVT, IEEE TETCI, Neural Networks, and Pattern Recognition.

\end{IEEEbiography}

\end{document}